\documentclass[11pt, a4paper, logo]{googlecloud}

\pdftrailerid{redacted}

\makeatletter
\renewcommand\bibentry[1]{\nocite{#1}{\frenchspacing\@nameuse{BR@r@#1\@extra@b@citeb}}}
\makeatother

\usepackage[authoryear, sort&compress, round]{natbib}

\usepackage[utf8]{inputenc} %
\usepackage[T1]{fontenc}    %
\usepackage{url}            %
\usepackage{booktabs}       %
\usepackage{nicefrac}       %
\usepackage{microtype}      %
\usepackage{amsmath}
\usepackage{graphicx}
\usepackage{multicol}
\usepackage{hyperref}       %
\usepackage[nameinlink]{cleveref}
\usepackage{bbm}
\usepackage{multirow}
\usepackage{makecell}
\usepackage{soul}
\usepackage{floatrow}
\usepackage{float}
\usepackage{wrapfig}
\usepackage{blindtext}
\usepackage{tablefootnote}
\usepackage{amsfonts}
\usepackage[flushleft]{threeparttable}
\usepackage{bbding}
\usepackage{xcolor}
\usepackage{xspace}
\usepackage{bm}
\usepackage{arydshln}
\usepackage{enumitem}
\usepackage{setspace}
\usepackage{color}
\usepackage{algorithm}
\usepackage{longtable}
\usepackage{tcolorbox}

\usepackage[normalem]{ulem}
\usepackage{ulem}
\usepackage[nomargin,inline,marginclue,draft]{fixme}
\usepackage{balance}
\usepackage{verbatim}
\usepackage{diagbox}
\usepackage{changepage}
\usepackage{amssymb}
\usepackage{pifont}
\usepackage{array}   %

\usepackage{kantlipsum, lipsum}
\usepackage{dsfont}

\usepackage{amsmath,amsfonts,bm}

\def\eqref#1{equation~\ref{#1}}

\def\1{\bm{1}}

\DeclareMathAlphabet{\mathsfit}{\encodingdefault}{\sfdefault}{m}{sl}
\SetMathAlphabet{\mathsfit}{bold}{\encodingdefault}{\sfdefault}{bx}{n}

\usepackage{hyperref}

\usepackage{algorithmic}
\usepackage{subcaption}
\usepackage{listings}

\newcommand{\rparagraph}[1]{\vspace{0.0mm}\noindent\textbf{#1}}

\definecolor{plotblue}{HTML}{7EA6E0}
\definecolor{plotpink}{HTML}{F19C99}
\definecolor{decomposed_vq}{HTML}{D79B00}
\definecolor{proposed_prompt}{HTML}{9673A6}
\definecolor{best_prompt}{HTML}{006633}

\definecolor{plotred}{HTML}{f77189}
\definecolor{plotgreen}{HTML}{33b07a}
\definecolor{plotpurple}{HTML}{cc7af4}
\definecolor{plotmagenta}{HTML}{f565cc}
\definecolor{plotazure}{HTML}{38a9c5}

\definecolor{codegreen}{rgb}{0,0.6,0}
\definecolor{codegray}{rgb}{0.5,0.5,0.5}
\definecolor{codepurple}{rgb}{0.58,0,0.82}
\definecolor{backcolour}{rgb}{0.95,0.95,0.92}
\lstdefinestyle{mystyle}{
    backgroundcolor=\color{backcolour},   
    commentstyle=\color{codegreen},
    keywordstyle=\color{magenta},
    numberstyle=\tiny\color{codegray},
    stringstyle=\color{codepurple},
    basicstyle=\ttfamily\scriptsize,
    breakatwhitespace=false,         
    breaklines=true,                 
    captionpos=b,                    
    keepspaces=true,                 
    numbers=left,                    
    numbersep=5pt,                  
    showspaces=false,                
    showstringspaces=false,
    showtabs=false,                  
    tabsize=2,
    frame=none,
    aboveskip=1pt,
    belowskip=1pt,
}
\definecolor{mainboxbg}{HTML}{F7F9FC}
\definecolor{mainboxborder}{HTML}{A1C6EA}
\definecolor{mainboxbg2}{HTML}{FFDEDE}
\definecolor{mainboxborder2}{HTML}{FF9393}

\newcommand{\ours}{\textcolor{black}{\textsc{Maestro}}\xspace} %

\title{\ours: Self-Improving Text-to-Image Generation via Agent Orchestration}

\correspondingauthor{\footnotesize{Corresponding authors: \{xingchenw, soarik\}@google.com}\\}

\renewcommand{\today}{}

\author[1]{Xingchen Wan}
\author[2*]{Han Zhou}
\author[1]{Ruoxi Sun}
\author[*]{Hootan Nakhost}
\author[1]{Ke Jiang}
\author[1]{Rajarishi Sinha}
\author[1]{Sercan Ö. Arık}

\affil[1]{Google}
\affil[2]{University of Cambridge}
\affil[*]{Work done at Google}

\begin{abstract}
Text-to-image (T2I) models, while offering immense creative potential, are highly reliant on human intervention, posing significant usability challenges that often necessitate manual, iterative prompt engineering over often underspecified prompts. This paper introduces \ours, a novel self-evolving image generation system that enables T2I models to autonomously self-improve generated images through iterative evolution of prompts, using only an initial prompt. \ours incorporates two key innovations: 1) self-critique, where specialized multimodal LLM (MLLM) agents act as `critics' to identify weaknesses in generated images, correct for under-specification, and provide interpretable edit signals, which are then integrated by a `verifier' agent while preserving user intent; and 2) self-evolution, utilizing MLLM-as-a-judge for head-to-head comparisons between iteratively generated images, eschewing problematic images, and evolving creative prompt candidates that align with user intents. Extensive experiments on complex T2I tasks using black-box models demonstrate that \ours significantly improves image quality over initial prompts and state-of-the-art automated methods, with effectiveness scaling with more advanced MLLM components. This work presents a robust, interpretable, and effective pathway towards self-improving T2I generation.
\end{abstract}

\begin{document}

\maketitle

\section{Introduction}
\label{sec:introduction}

The advent of powerful text-to-image (T2I) models, including systems like Imagen~\citep{saharia2022photorealistic, baldridge2024imagen}, DALL-E~\citep{ramesh2022hierarchical, betker2023improving}, and the integrated multimodal generation capabilities in frontier models like Gemini 2.0~\citep{reid2024gemini} and GPT-4o~\citep{hurst2024gpt}, has significantly transformed the landscape of digital content creation. These models empower users to generate intricate and often stunning visuals directly from textual descriptions, offering unprecedented creative potential. Central to their operation is the text prompt, which serves as the primary (and for \textit{black-box} models, the \textit{only}) mechanism for controlling the generation process. Consequently, T2I models are inherently highly sensitive to this input prompt.

This sensitivity often extends beyond the intended level of control, presenting significant usability challenges. Generated images can vary drastically even when prompts are semantically equivalent, and models exhibit differing responses to various prompting styles. For instance, recent findings suggest a preference for detailed, descriptive prompts over abstract concepts that necessitate implicit reasoning~\citep{niu2025wise}. This challenge is compounded by the inherent representational gap between a user's conceptual intent, its textual articulation in a prompt, and the resulting image: for example, users are found to frequently \textit{under-specify} their desired output, providing prompts that lack the necessary detail or nuance for the model to generate optimal results~\citep{hahn2024proactive}. As a consequence of this combined sensitivity and under-specification, users typically resort to a manual, iterative process of prompt refinement. They generate an image, evaluate its shortcomings relative to their goal, modify the prompt, and repeat the cycle until a satisfactory output is achieved. This trial-and-error approach is not only costly and time-consuming, but also demands considerable expertise in prompt engineering, hindering broader accessibility and efficient use of these powerful tools.

Some of the aforementioned challenges are not limited to T2I models -- for example, the prompt sensitivity in textual-output large language models (LLMs) has seen extensively studied~\citep{zhao2021calibrate, zhou2024batch}, and \textit{automated prompt optimization} (APO) techniques, which often employ an LLM to iteratively propose improved prompt candidates that maximize an user-specified objective function on a provided labeled validation set, have been developed and widely adopted~\citep{zhou2022large}. However, generalizing this iterative refinement process for multimodal T2I generation is less straightforward. A core difficulty lies in \textit{evaluation}: APO frames prompt engineering as optimization, which inherently requires an \textit{optimization objective}. However, assessing the quality of generated images is inherently subjective, multifaceted (encompassing fidelity to the prompt, aesthetic appeal, coherence, style consistency, etc.), and lacks objective ground-truth references. Previous attempts to quantify quality using image reward models~\citep{xu2023imagereward} or employing LLMs to decompose prompts for proxy optimization~\citep{cho2023davidsonian, hu2023tifa} have shown limitations. These proxy metrics often struggle to fully reflect the nuances of multimodal evaluations (e.g., they may only aim to represent one aspect of quality like prompt-image consistency but others like the overall aesthetic) and, therefore, often exhibit weak correlation with human perceptual judgments~\citep{ross2024makes}. Other approaches were proposed to directly use the multimodal understanding of multimodal LLMs (MLLMs) to guide generation~\citep{liu2024language}. However, a crucial issue is that without a clearly defined objective function, even identifying the best out of many generations to return to the users or setting a termination criterion can be challenging. Lastly, methods have also been proposed to tune models via offline training aiming to automatically improve user prompts~\citep{hao2023optimizing, datta2023prompt}. These, however, can be bottlenecked by the intensive data annotation cost, training data coverage (i.e., poor generalization to new tasks and models), and the need to host a bespoke model simply for prompt refinement. 

In this paper, we introduce \ours, a novel agentic system specifically designed to address these challenges and allow T2I generations to \textit{self-improve} via prompt refinement from carefully orchestrating black-box MLLM and T2I model calls. Our approach mimics the iterative refinement strategy (similar to how humans would approach) but achieves self-improvement completely autonomously, requiring only an initial user prompt input. Key to our system are two innovations:
\begin{itemize}
    \item \textit{Self-improving multimodal generation from multi-agentic critique}: We structure our system with multiple specialized agents. Inspired by how humans identify and address specific unsatisfactory aspects in previous attempts, we utilize MLLMs as `critics'. Crucially, unlike prior works that used MLLMs to assist generation primarily as \textit{discriminating reward models} to generate scalar reward scores~\citep{manas2024improving}, we directly take advantage of their sophisticated {multimodal understanding and reasoning} capabilities akin to \textit{generative reward models}~\citep{mahan2024generative} to generate edit signals instead. These critic agents analyze generated images conditioned on the user prompt to identify specific weaknesses, correct for under-specification, and generate targeted, interpretable signals for prompt editing. A separate LLM agent then acts as a `verifier', integrating these signals while ensuring the revisions remain grounded in the user's original intent.
    \item \textit{Pairwise objective}: Acknowledging the subjective and multifaceted nature of evaluations for visual data, where different aspects like prompt consistency, fidelity, and aesthetics hold varying importance for different users or tasks, we eschew single-score, point-wise objectives. Instead, we adopt a pairwise comparison objective, a methodology well-established in fields such as reinforcement learning with human feedback (RLHF)~\citep{ziegler2019fine, christiano2017deep, ouyang2022training} for navigating similarly ambiguous problems like aligning models with human preferences. At each iteration, our system utilizes an MLLM-as-a-judge to conduct a \textit{binary tournament} consisting of a series of head-to-head comparisons between the image generated with the current prompt and the best image generated so far. This iterative comparison process continues until a predefined budget (e.g., number of iterations) or another patience criterion is met, ultimately returning the incumbent best generation.
\end{itemize}
We validate our approach through extensive experiments on high-complexity T2I generation tasks using black-box models. Our results show significant improvements in image quality compared to both the initial user prompts and prompts generated by state-of-the-art automated methods, confirmed by both automatic metrics and human evaluations. Notably, we also demonstrate that the effectiveness of our pipeline scales with the capabilities of its components, achieving further performance gains when employing more advanced MLLMs, such as Gemini 2.0, within the multi-agent framework. This work presents a robust, interpretable, and effective automated solution for enhancing T2I generation by intelligently navigating the complexities of prompt sensitivity and subjective evaluation.

\section{Related Work}
\label{sec:analysis}
In this section, we provide an overview of the state-of-the-art in APO for text-to-image (T2I) generation, and we broadly categorize existing approaches into \textit{train-time} and \textit{test-time} methods. Our focus is on techniques applicable to T2I models as black boxes (i.e., prediction API-only access), and thus we exclude approaches that necessitate direct modifications, depend on certain model architecture, or other white-box accesses~\citep[\textit{inter alia}]{fan2023dpok}.

\rparagraph{Train-time approaches.} Train-time approaches aim to directly learn a mapping from a user's initial prompt to a more effective prompt via calling a dedicated model, often a fine-tuned LLM (referred to as the \textit{optimizer model}), in a one-shot manner. A seminal work in this area is \textit{Promptist}~\citep{hao2023optimizing}, which collected a dataset of initial user prompts paired with corresponding optimized prompts. An LLM was then trained using supervised fine-tuning (SFT) and RLHF to learn the transformation. Other works like have expanded in this direction by, for example, enhancing diversity \citep{datta2023prompt, yun2025learning} and improving finer-grained control~\citep{mo2024dynamic}. Instead of relying on human annotation, other works~\citep{cao2023beautifulprompt} have reused existing large-scale databases like DiffusionDB~\citep{wang2022diffusiondb} with rule-based filtering for dataset construction for subsequent SFT-preference learning pipeline with automated image quality metrics (e.g., PickScore~\citep{kirstain2023pick}). Overall, while these train-time methods effectively amortize the computational cost of optimization to the training phase, they share several limitations. A primary drawback is their near-universal dependence on large-scale datasets, the manual creation of which demands significant labor and specialized expertise. While automated pipelines are available, they typically rely on proprietary T2I systems to generate reference images and/or rely on specific reward models (as in T2I, ground-truth ``optimal'' prompts and images are ill-defined), potentially making the data and subsequently, the tuned optimizer LLM, specific to the generative model and data used. Furthermore, while optimizing prompts via a dedicated tuned model has key advantages like low-latency, it is challenging to incorporate feedback from the actual generated image at test time, unlike how humans iteratively refine prompts based on visual results. The fact that a bespoke model has to be trained and served also means it is difficult to leverage rapid advances in foundational T2I and MLLM models alike, as the optimizer model would need retraining with newly generated datasets to stay current.

\begin{figure}[t!]
        \centering
        \includegraphics[width=\linewidth, trim={0.7cm 0cm 0.7cm 0cm}, clip]{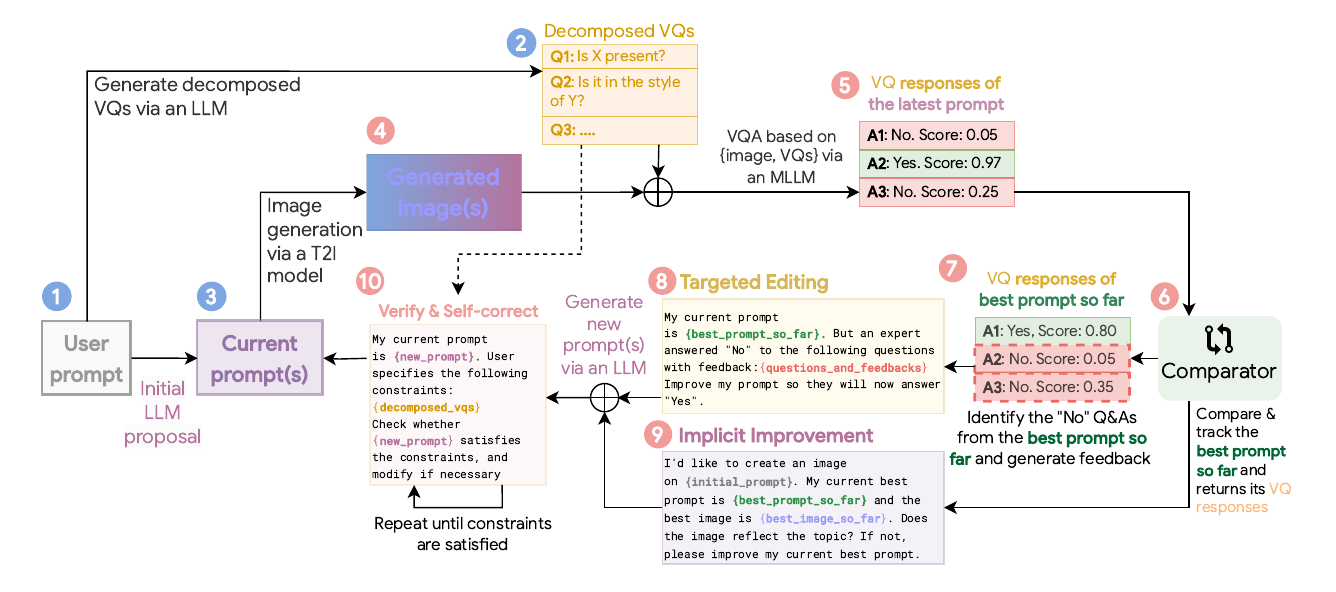}
            \caption{The overall pipeline of \ours: \textbf{1) \textcolor{plotblue}{At initialization}}, we use the \textcolor{gray}{user prompt} ({Block 1}) \textbf{a)} with an existing method like DSG~\citep{cho2023davidsonian} to generate \textcolor{decomposed_vq}{\textit{decomposed visual questions} (DVQs)} ({Block 2}), where each question reflects a property/characteristic desired by the user in the generated image, and \textbf{b)} we obtain an \textit{\textcolor{proposed_prompt}{initial LLM proposal}}, an LLM-generated prompt that is the revised version of the user prompt taking into account of the generic instructions on T2I prompting best practices ({Block 3}). \textbf{2) \textcolor{plotpink}{During each iteration}}, \ours generates an image with the current prompt by calling the T2I model ({Block 4}), scores the image against the \textcolor{decomposed_vq}{DVQs} ({Block 5}), and uses the Comparator ({Block 6}) to determine the \textcolor{best_prompt}{\textit{best prompt and image so far}}. We then perform both \textbf{a)} targeted editing on the \textcolor{best_prompt}{best prompt so far} by asking an MLLM to generate critique and improvement signals based on responses to the \textcolor{decomposed_vq}{DVQs}, each of which represents a \textit{specific} inadequacy of the generated image ({Blocks 7-8}) against the user prompt and \textbf{b)} generally asking an MLLM to reflect on the prompt and image and proposing improvements ({Block 9}). We then collect these generated prompts and verify them against the \textcolor{decomposed_vq}{DVQs} to ensure the revised prompts remain grounded on the user intent (Block 10), and finally return the \textcolor{proposed_prompt}{prompt proposals} for the next iteration. \textbf{3) At termination} (i.e., budget/patience exhaustion), we simply return the \textcolor{best_prompt}{{best prompt and image so far}} maintained by the Comparator ({Block 6}) to the user.
            }
        \label{fig:mainfig}
    \end{figure}

\rparagraph{Test-time approaches.} In contrast to train-time methods, test-time approaches perform optimization during the inference phase. These methods typically involve an iterative loop of generating an image, evaluating it based on some feedback signal, and revising the prompt accordingly. This process more closely mimics the iterative refinement strategy employed by humans and is common in APO in textual generative models. Specifically, some test-time methods rely on explicit, quantitative, objective functions: OPT2I~\citep{manas2024improving} uses an improved Davidsonian Scene Graph (DSG)~\citep{cho2023davidsonian} score and CLIPScore~\citep{hessel2021clipscore} as optimization objectives; an LLM was employed to rewrite or paraphrase the prompt in an OPRO-style~\citep{yang2023large} loop, where the optimizer proposes prompts predicted to achieve higher scores based on the history of prompts and their associated scores. \citet{wang2024discrete} mainly relies on word substitutes (e.g., antonyms and synonyms), and use evolutionary algorithm to search against the CLIP loss. Other approaches forgo explicit numerical objectives and directly utilize the capabilities of MLLMs. For example, \citet{liu2024language} iteratively query an MLLM by providing it with the initial user prompt, the most recent prompt, and the corresponding generated image. An MLLM is then asked to assess whether the image aligns with the initial intent and, if not, to suggest specific modifications to the last prompt. Test-time methods offer distinct advantages: utilizing more extensive test-time compute, they are generally model-agnostic regarding the underlying T2I generator and require no dedicated training phase or large-scale data collection, thus eliminating associated costs. This inherent flexibility allows them to readily incorporate and benefit from advancements in the LLMs or MLLMs used for feedback and revision as more powerful models become available. However, existing iterative optimization methods often tend to generate new prompts by paraphrasing the user prompts, which may not effectively address the core issue of user prompt under-specification (since the generated prompts are still paraphrased versions of the under-specified user prompt). While the idea of MLLM-guided optimization has been explored, a significant limitation is that prior works employing MLLMs have largely used them as black-box reward models to produce a scalar quality score (e.g., DSG Score) but did not fully leverage their sophisticated multimodal reasoning capabilities to generate interpretable and targeted feedback for revision. Furthermore, approaches that do use MLLMs for critique but lack an explicit objective function often lack a clear mechanism for maintaining the best-found candidate across iterations, typically defaulting to returning the generation from the final step; as we will show in Sec.~\ref{sec:methods}, \ours uses pairwise objective and rich feedback to address these issues.

\section{Methodology}
\label{sec:methods}

\begin{minipage}{0.48\textwidth}
\vspace{-5mm}
\begin{algorithm}[H]
\begin{footnotesize}
	    \caption{\ours.}
	    \label{alg:main_alg}
	\begin{algorithmic}[1]
		\STATE \textbf{Input}: User prompt $p_u$, T2I model $\texttt{T2I}(\cdot)$, optimizer LLM $\texttt{LLM}(\cdot)$, judge MLLM $\texttt{MLLM}(\cdot)$ (can be the same as $\texttt{LLM}$ if it handles multimodal inputs), max number of rounds $T$, (optionally) early-stop patience $m$.
	\FOR{$t = \{0, ..., T\}$}
	\IF{$t=0$}
	\STATE  \textbf{(\S\ref{subsec:init})} Generate DVQs $Q \leftarrow \texttt{DVQGen}_{\text{LLM}}(p_u)$ \& initial LLM proposal $p^{(0)} \leftarrow \texttt{Init}_{\text{LLM}}(p_u)$ and set $\{p^*_{:t}, I^*_{:t}\} \leftarrow \texttt{None}$.
	\ELSE
	\STATE \textbf{(\S\ref{subsec:new_prompt})} Generate new prompt proposals with targeted editing and/or implicit improvements: $p^{(t)} \leftarrow \texttt{TargetedEdit}_{\text{LLM}}(p^*_{:t-1}, Q, R^*_{:t-1}) \cup \texttt{ImpImprove}_{\text{MLLM}}(p^*_{:t-1}, p_u, I^*_{:t-1})$ 
	\ENDIF
	\STATE \textbf{(\S\ref{subsec:intent_grounding})} Refine and self-correct prompt proposals to ensure grounding on original user intent: $p^{(t)} \leftarrow \texttt{Refine}_{\texttt{LLM}}(p^{(t)}, Q)$.
	\STATE Generate image $I_t \leftarrow \text{T2I}(p^{(t)})$ and obtain DVQ responses: $R^{(t)} = \texttt{VQA}_{\text{MLLM}}(I^{(t)}, Q)$.
	\STATE \textbf{(\S\ref{subsec:pairwise})} Update best prompt/image so far: $\{p^*_{:t}, I^*_{:t}\} \leftarrow \{p^{(t)}, I^{(t)}$ if $t = 0$;  otherwise $\{p^*_{:t}, I^*_{:t}\} \leftarrow \texttt{Compare}_{\text{MLLM}} \big(\{p^{(t)}, I^{(t)}\}, \{p^*_{:t-1}, I^*_{:t-1}\} \big)$.
	\ENDFOR
	\RETURN Best prompt/image at termination $\{p^*, I^*\}$ after $T$ steps or $m$ non-improving steps.
	\end{algorithmic}
\end{footnotesize}
\end{algorithm}
\end{minipage}
\hfill
\begin{minipage}{0.5\textwidth}
\centering
\begin{scriptsize}
\resizebox{\textwidth}{!}{
\begin{tabular}{m{8cm}}
\hline
\textbf{User prompt $p_u$}: {studio close-up shot of an antique book with  knowledge is power  painted in gold on the cover in thick flowing brushed calligraphy} \\
\hline
\textbf{Initial LLM proposal} $p^{(0)}$: {Studio close-up photograph of an antique leather-bound book lying open on a polished mahogany table. The cover, illuminated by a soft Rembrandt-esque light from the upper left, displays the phrase "Knowledge is Power" painted in thick, flowing, gold leaf calligraphy reminiscent of medieval illuminated manuscripts. The edges of the book are worn and slightly frayed, hinting at age and frequent use. Dust motes dance in the beam of light, adding a touch of atmospheric depth. The background is a slightly out-of-focus dark oak bookshelf filled with similarly aged tomes, creating a sense of scholarly reverence.} \\
\hline
\vspace{2mm}
\centering
\includegraphics[width=0.48\linewidth]{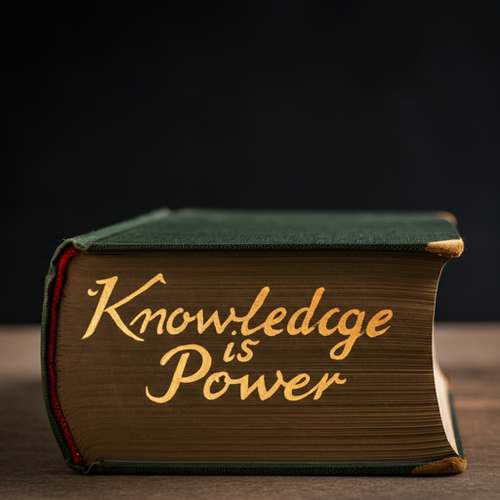}
\includegraphics[width=0.48\linewidth]{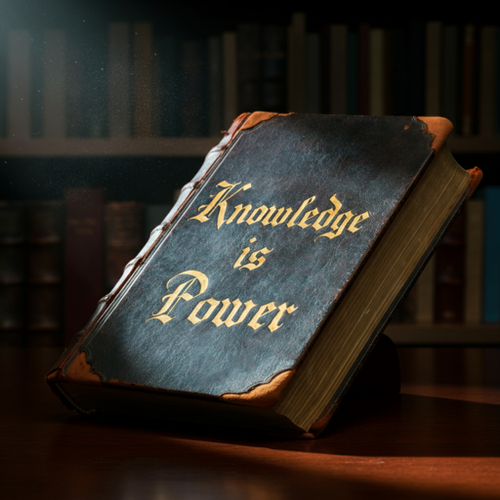}
\end{tabular}
}
\captionof{figure}{Illustrating examples of an user prompt $p_u$, the corresponding initial LLM proposal prompt $p^{(0)}$ and their corresponding generations using Imagen 3 (\textit{left}: $p_u$; \textit{right}: $p^{(0)}$. Refer to additional illustrations in App.~\ref{app:additional_qualitative}.}
\label{fig:init_rewrite}
\end{scriptsize}
\end{minipage}

In this section, we present \ours (standing for \underline{M}ulti-\underline{A}gent \underline{Orch}estration and highlighting it as a \textit{maestro} for image generation), our proposed framework for automated prompt optimization in T2I generation. The high-level workflow of \ours is illustrated in Fig.~\ref{fig:mainfig}, and a detailed pseudo-code representation is provided in Algorithm~\ref{alg:main_alg}. Following the taxonomy discussed in \S\ref{sec:analysis}, \ours primarily adopts a test-time optimization framework. This design choice circumvents the need for large-scale labeled training data and ensures that our method is model-agnostic, capable of working with various black-box T2I generators and (M)LLM optimizers. While test-time iteration inherently increases runtime compared to one-shot generation that might be utilizing train-time approaches, it mirrors the iterative refinement process users often undertake manually to achieve satisfactory results, but it is worth emphasizing that \textbf{\ours features completely autonomous self-improvement} and, by default, requires only initial user prompt as the manual component. Furthermore, as we will discuss below, our framework is designed to seamlessly integrate and benefit from train-time approaches by potentially using their output during its initialization phase, thereby unifying the strengths of both paradigms. While we provide a high-level walkthrough in the captions of Fig.~\ref{fig:mainfig}, we discuss each key component of \ours in detail below.

\subsection{Initialization (Blocks 1-3 in Fig.~\ref{fig:mainfig})}
\label{subsec:init}
Given an initial user prompt $p_u$, \ours performs two key operations once at the beginning of the process to establish a robust starting point and necessary components for subsequent iterations:

\rparagraph{Generating initial LLM proposal} (Block 3): As discussed in \S\ref{sec:introduction}, user prompts often lack the structure or clarity that facilitates optimal T2I generation. Recognizing that some improvements can be achieved by aligning the prompt with general prompt engineering best practices, we first employ a text-LLM (Gemini 1.5 Pro in our experiments) guided by a meta-prompt containing established guidelines (e.g., focusing on clear subject description, defining context and scene details, specifying artistic style; see App.~\ref{app:details} for details) to transform the raw user prompt $p_u$ into an initial, potentially more effective prompt $p^{(0)}$. The objective here is twofold: 1) to achieve a better starting point for the iterative process, potentially reducing the number of iterations needed, and 2) to provide a natural bridge between train-time and test-time methods: while we currently use a general instruction-following LLM, this step could readily incorporate a specialized prompt optimization model fine-tuned via a train-time approach such as one outlined in \S\ref{sec:analysis}, if available. We show an illustrative example in Fig.~\ref{fig:init_rewrite} where we compare $p_u$ against $p^{(0)}$ and their corresponding generated images, and we find that while simplistic, the initial LLM rewrite is already surprisingly effective in enriching the text prompt with descriptive information and in enhancing the resultant generation (e.g., compared to the image generated by $p_u$, the image generated by $p^{(0)}$ features text that is correctly painted on the cover, and the book has a better antique appearance), although there is still room for improvement (e.g., the font of the phrase is still not in `thick flowing brushed calligraphy' requested by the user).

\rparagraph{Generating decomposed visual questions (DVQs)} (Block 2): Concurrently, we leverage techniques previously developed for T2I evaluation to decompose the initial user prompt $p_u$ into a set of DVQs, each probing a desirable property implied by the prompt via an LLM call~\citep[inter alia]{cho2023davidsonian, hu2023tifa, yarom2023you}. Specifically, we use a pipeline inspired by \citet{cho2023davidsonian} but instead of relying on in-context learning only, we additionally prompt the LLM once more to refine on the initial questions to improve coherence and precision. For example, the user prompt in Fig.~\ref{fig:init_rewrite} yields the following DVQs ($Q \leftarrow \texttt{DVQGen}_{\text{LLM}}(p_u)$, where ${Q} = \{q_1, ..., q_n\}$ is the set of DVQs and $\texttt{DVQGen}_{\text{LLM}}(\cdot)$ refers to the operation of generating DVQs by prompting an LLM with $p_u$):

\begin{footnotesize}
\begin{enumerate}
    \item Is the image a close-up of the book?
    \item Is there a book?
    \item Is the book antique?
  \item Does the cover appear to be part of an antique book? (e.g., material, wear, binding style)
  \item Does the cover say ``knowledge is power''?
  \item Is the ``knowledge is power'' text painted in gold?
  \item Does the ``knowledge is power'' text appear to be made with thick brushstrokes?
  \item Does the ``knowledge is power'' text have a flowing calligraphic style?
  \item Does the lighting appear to be studio quality (e.g., even, controlled)?
  \item Is the background uncluttered and consistent with a studio setting?
\end{enumerate}
\end{footnotesize}

These DVQs are reused at multiple stages within our pipeline, but importantly, unlike the aforementioned work that primarily uses the responses to these questions to evaluate prompt-image consistency, we do \textit{not} use the scalar, aggregated answer correctness of these DVQs as the optimization objective. While convenient, simple aggregation (e.g., summing or averaging binary scores) dilutes the rich, multifaceted information captured by the individual questions and struggles to accurately represent complex user preferences, often showing poor correlation with human judgments~\citep{ross2024makes}. This also deviates from how humans typically form preferences through comparative assessment. Instead, \ours optimizes directly based on pairwise preferences generated by an LLM judge, as detailed next.

\subsection{Pairwise Objective and Comparator (Block 6)} 
\label{subsec:pairwise}
At $t = \{0, 1, ...\}$-th iteration with the current prompt proposal $p^{(t)}$ (when $t=0$, it is the initial LLM proposal mentioned above; for $t>0$, it is generated via the techniques to be outlined in \S\ref{subsec:new_prompt}), we generate corresponding image(s) by calling the T2I model: $I^{(t)} \leftarrow \texttt{T2I}(p^{(t)})$ (Block 4). We then follow the standard VQA pipeline to evaluate $I^{(t)}$ against the list of DVQs generated during initialization to obtain the responses with an MLLM: $R^{(t)} \leftarrow \{r_i := \texttt{VQA}_{\text{MLLM}}(I^{(t)}, q_i) \, \forall q_i \in Q\}$ where we follow \citet{ren2023self} to define the response as the probability of the ``Yes'' token given the image and the DVQs $r_i^{(t)} := p_{\texttt{MLLM}}(\text{``Yes''}|I^{(t)}, q_i) \in [0, 1]$ and store the 3-tuple $\{p^{(t)}, I^{(t)}, R^{(t)}\}$ (Block 5) in \textbf{Comparator} (Block 6), a core component of \ours, where we adopt the \textit{pairwise preference from the LLM feedback} to track the \textit{best prompt \& image up to the $t$-th iteration} (denoted as $p^*_{:t}$ and $I^*_{:t}$, respectively).

\begin{wrapfigure}{l}{0.45\textwidth}
\centering
\includegraphics[width=\linewidth, trim={1.1cm 0.3cm 1cm 0.4cm}, clip]{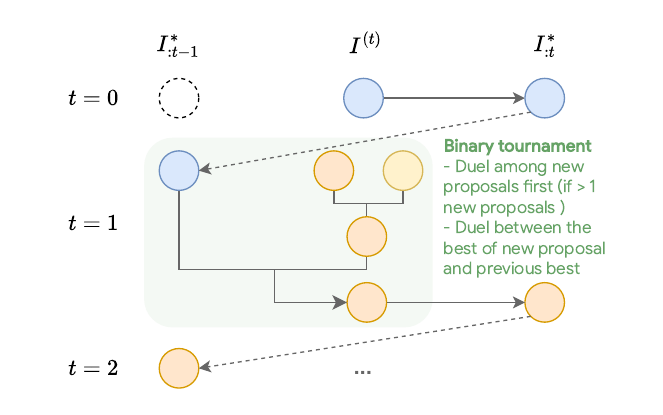}
\caption{Illustration of the binary tournament performed by Comparator (Block 6) to obtain the best image so far at $t$-th iteration $I^*_{:t}$ given new proposal(s) $I^{(t)}$ and the previous best $I^*_{:t-1}$.}
\label{fig:binary_tournament}
\end{wrapfigure}

\rparagraph{Rationale.} As mentioned, evaluating generative images is inherently subjective, multifaceted and is without a reference ground-truth, which we argue closely parallels the problem addressed by RLHF and reinforcement learning with AI feedback (RLAIF, \citet{lee2023rlaif}), where human/AI-annotated pairwise preferences are used to effectively align models with complex, hard-to-quantify human values or instructions -- goals where assigning a consistent, absolute scoring function is notoriously difficult. However, whereas RLHF/RLAIF typically trains a reward model to ultimately approximate the pairwise preference to a scalar value through, for example, learning a Bradley-Terry model~\citep{bradley1952rank}, we propose to optimize against the pairwise objective directly as elicited by the MLLM-as-a-judge in iteration, bypassing the potential inaccuracies, biases and limitations introduced by an intermediate reward model approximation through a procedure we outline below.

\rparagraph{Procedure.} Directly optimizing against a pairwise objective requires a reference for comparison. At $t$-th iteration, the Comparator ($\texttt{Compare}_{\text{MLLM}}(\cdot)$) determines the best prompt/generation by comparing the latest candidate prompt/generation against the \textit{incumbent} best: 
\begin{equation}
    \{p^*_{:t}, I^*_{:t}\} \leftarrow \texttt{Compare}_{\text{MLLM}} \Big(\{p^{(t)}, I^{(t)}\}, \{p^*_{:t-1}, I^*_{:t-1}\} \Big).
\end{equation}
To initialize, we assign $p^*_{:0} \leftarrow p^{(0)}, I^*_{:0} \leftarrow I^{(0)}$ (the $t=0$ row in Fig.~\ref{fig:binary_tournament}). In subsequent iterations $t > 0$,  we conduct a side-by-side ``duel'' between $I^{(t)}$ and $I^*_{:t-1}$, conditional on the \textit{user prompt} $p_u$ with a capable MLLM (we use Gemini 1.5 Pro in our experiments) prompted as follows:

\begin{tcolorbox}[colback=mainboxbg, colframe=mainboxborder, coltitle=black, fonttitle=\bfseries, title=MLLM-as-a-judge prompt for binary tournament, boxrule=0.5mm, arc=2mm, outer arc=2mm]
\footnotesize

You are an expert in critiquing images. Given a user prompt, a text-to-image model generated two images (``Image A'' and ``Image B''), and your task is to pick the overall better image. The user prompt is:

``\{ \texttt{user\_prompt} \}''

Image A: 
\{ \texttt{image\_A} \}

Image B: 
\{ \texttt{image\_B} \}

Please carefully explain the rationale \textbf{before} giving your final choice. Make sure you finish your answer with ``\texttt{<answer> X </answer>}'', where X is ``A'' or ``B''.
\end{tcolorbox}
As illustrated in Fig.~\ref{fig:binary_tournament}, if more than one image is generated in one iteration, we conduct a tournament using this pairwise comparison to identify the single best candidate among them before comparing it against $I^*_{:t}$. To improve the robustness of the comparison and in particular, to reduce \textit{position bias} inherent in LLMs~\citep{shi2024judging, chiang2023can, zheng2023large, zhou-etal-2024-fairer}, we query the MLLM judge $2n$ times (with $I^{(t)}$ and $I^*_{:t-1}$ each appearing as \texttt{image\_A} $n$ times) in parallel with temperature of 0.7 and break ties randomly to select the final winner, and finally assign $I^*_{:t} \leftarrow I^{(t)}$ if $I^{(t)}$ wins the duel or $I^*_{:t} \leftarrow I^*_{:t-1}$ otherwise. This iterative process of proposal, generation, and pairwise comparison continues until a predefined budget (e.g., maximum number of T2I calls $T$) is exhausted, or optionally, if $\{p^*_{:t}, I^*_{:t}\}$ do not change for $m$ consecutive iterations (patience criterion). The comparator then returns the final results $\{p^*, I^*\}$ to the users.

\subsection{Generating New Prompt Proposals (Blocks 7-9)}
\label{subsec:new_prompt}

The mechanism by which new prompts are proposed based on prior evaluations lies at the heart of any iterative APO algorithm: \ours employs a dual-generator strategy, leveraging two distinct approaches to create candidate prompts for the next iteration upon the best prompt-image pairs so far $\{p^*_{:t}, I^*_{:t}\}$. This design acknowledges the multifaceted nature of T2I evaluation and allows us to capitalize on different types of feedback signals. The two generators are designed to be complementary: the first, which we term ``\textit{targeted editing}'', explicitly targets specific defects identified via structured feedback on previous generations, while the second, which we term ``\textit{implicit improvement}'', aims for holistic improvements based on MLLM judge's general multimodal understanding capabilities.

\begin{table}[t!]
\centering
\resizebox{\textwidth}{!}{
\begin{tabular}{m{7cm} m{9cm} m{7cm}}
\toprule
Current image $I^*_{:t-1}$ & Selected DVQ(s), rationalization \& suggestions & New prompt \& generation $\{p^{(t)}, I^{(t)}\}$ \\
\cmidrule(lr){1-1} \cmidrule(lr){2-2} \cmidrule(lr){3-3}
\includegraphics[width=.8\linewidth]{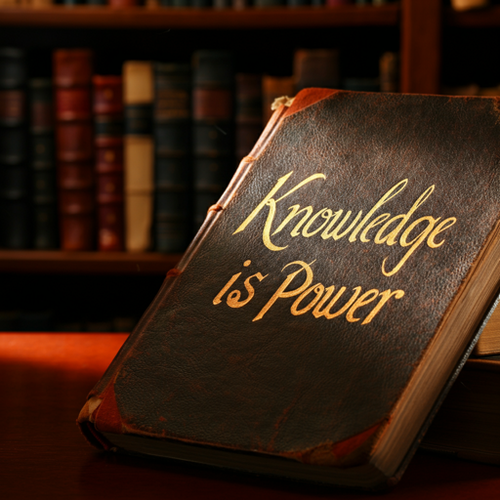}
& \textit{Question}: Does the "knowledge is power" text appear to be made with thick brushstrokes? \newline \textit{Rationalization}:  {The reviewer likely answered "No" \textbf{because the "Knowledge is Power" text appears to be written with a relatively thin, elegant script, rather than thick brushstrokes}.  While the letters are stylized, they maintain a smooth, flowing quality, resembling calligraphy or penmanship more than the textured, impasto effect of thick paint applied with a brush...} \newline \textit{Suggestion}: ... the text should be re-rendered to appear as if created with thick brushstrokes. This could be achieved by \textbf{adding texture to the letters, making the edges less precise and more irregular, and incorporating variations in the gold color's density to simulate the uneven application of paint}...
& \includegraphics[width=.8\linewidth]{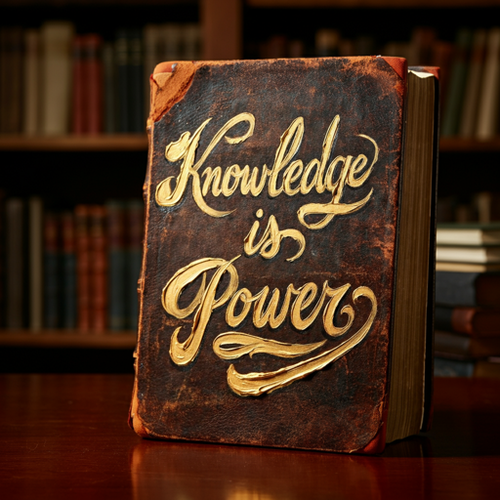} \newline ... features the phrase "Knowledge is Power" painted in \textbf{thick, impasto gold calligraphy, with visible brushstrokes, ridges, and uneven texture, as if applied with a thick brush} ...\\
\cmidrule(lr){1-1} \cmidrule(lr){2-2} \cmidrule(lr){3-3}

\includegraphics[width=.8\linewidth]{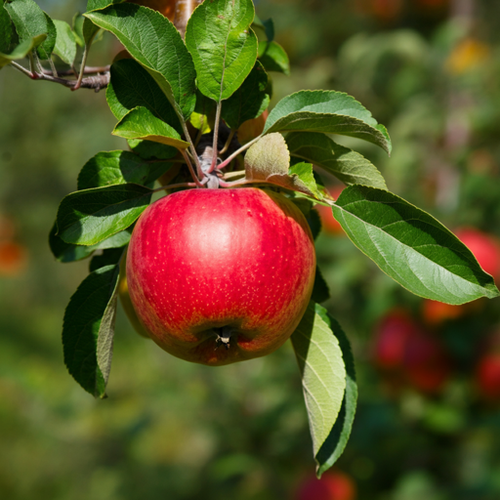} 
& \textit{Questions}: Are the apples square? / Are the leaves circular? \newline \textit{Rationalization}: The reviewer answered "No" because \textbf{the question specifically asks about a square shape, and the provided image depicts a typical, naturally round apple} / The reviewer answered "No" because the \textbf{leaves on the apple tree are ovate or elliptical, not circular.}... \newline \textit{Suggestion}:  * To satisfy the prompt and elicit a "Yes" response, the apple needs to be depicted as square. \textbf{This could be achieved by generating an image of a cuboid apple hanging from a branch}... / To create an image where the leaves are circular, the leaf shape needs to be fundamentally altered. \textbf{The prompt should include phrases like "perfectly round leaves", "circular foliage", or "leaves shaped like discs", ... Using terms like "coin-shaped leaves" can further reinforce the desired circularity.}
& \includegraphics[width=.8\linewidth]{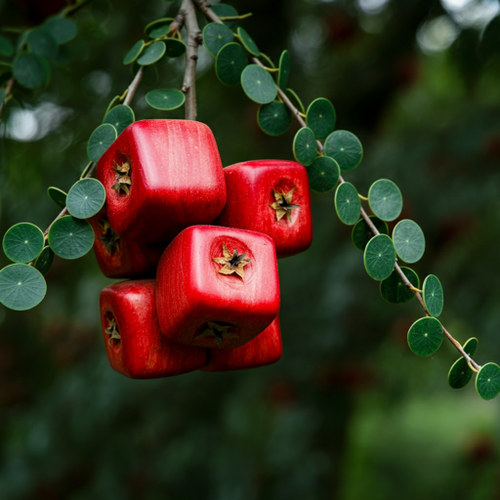} \newline Square, visibly seamed red apples, \textbf{carved into cuboid shapes}, hanging from a tree with \textbf{perfectly round, disc-shaped, coin-shaped green leaves. Circular foliage. Leaves shaped like discs.}\\
\bottomrule
\end{tabular}
\vspace{-2mm}
\captionof{figure}{Illustrating examples of the targeted editing pipeline (\S\ref{subsec:new_prompt}) with 1) current best image in the step \textit{before} applying targeted editing; 2) selected DVQ(s) where the judge model answered ``No'' and excerpts of LLM-generated rationalization and suggestions; and 3) the prompt excerpt and generated image \textit{after} incorporating the suggestions (refer to App.~\ref{app:additional_qualitative} for the full prompts). Key parts of {rationalization}, {suggestion} and resulting{revised prompt} are \textbf{bolded}. The initial user prompts $p_u$ are \textit{studio close-up shot of an antique book with  knowledge is power  painted in gold on the cover in thick flowing brushed calligraphy} (\textit{top row}) and \textit{square red apples on a tree with circular green leaves} (\textit{bottom row}), respectively. Figure is better viewed in color.}
\label{fig:targeted_editing}
}
\end{table}

\rparagraph{Targeted editing (Blocks 7-8).} Aiming to mimic the human process of identifying \textit{specific} deficiencies in a generated image and making targeted adjustments to the prompt to correct them, this generator reuses the DVQs generated at initialization $Q$ and the MLLM's responses to them for the best image so far $R^*_{:t-1}$. In contrast to prior works like \citet{manas2024improving} that primarily used the binary ``Yes/No'' responses and their aggregated scores as the optimization objective, we propose to use them \textit{beyond} numerical feedback by noticing two key observations: First, we note that a key advantage of VQA-based scoring is their inherent \textit{interpretability}, where each question $q_i \in Q$, such as the example given in \S\ref{subsec:init}, corresponds to a specific, desired characteristic that is human-interpretable. Second, a key forte of modern (M)LLMs is their \textit{reasoning capabilities} in explaining \textit{why} they made a certain response -- such a textual, non-numerical feedback has shown to be useful both to improve the usefulness of the eventual numerical output as a reward signal~\citep{mahan2024generative} and, as \textit{textual gradients}, to outperform numerical feedback-only APO in the textual tasks~\citep{pryzant2023automatic, yuksekgonul2024textgrad}. To capitalize on these observations, we \textbf{1)} identify a subset of DVQs where the MLLM judge provided a ``No'' answer (i.e., $r^*_{i, :t-1} = p_{\texttt{MLLM}}(\text{``Yes''}|I^*_{:t-1}, q_i) < 0.5$, indicating $I^*_{:t-1}$ failed to meet a criterion); \textbf{2)} prompt the MLLM to generate a \textit{textual rationalization} explaining \textit{why} the image failed according to that question to provide an interpretable insight into the failure mode; and \textbf{3)} conditional on the rationalization, a specific edit or rewrite to the prompt aimed \textit{precisely} at rectifying the identified shortcoming -- we illustrate this process with an example in Fig.~\ref{fig:targeted_editing} (with additional qualitative examples in App.~\ref{app:additional_qualitative}) and show the detailed prompt to achieve the above in App.~\ref{app:details}.

\rparagraph{Implicit Improvement (Block 9).} Complementary to the \textit{explicit} targeted editing approach above where we target specific inadequacies, we also employ a second generator to focus on \textit{implicit} holistic enhancement where we prompt a powerful MLLM to assess the current best image ($I^*_{:t-1}$) more broadly in the context of the prompts ($p^*_{:t-1}$ and $p_u$) and to provide prompt improvement without being strictly tied to the predefined DVQs. To achieve so, we incorporate the prompting subroutine introduced in \citet{liu2024language} to leverage the MLLM optimizer for general visual critique -- but a key distinction is that we apply it always to the best image so far $I^*_{:t-1}$ instead of the last generation $I^{(t-1)}$ in \citet{liu2024language}, given that our comparator now explicitly tracks the optimization progress. We again refer the readers to App.~\ref{app:details} for the detailed prompt used.

\subsection{User Intent Grounding via Self-Verification (Block 10)}
\label{subsec:intent_grounding}

One noticeable differentiator of our prompt generators in \S\ref{subsec:new_prompt} is the encouragement of \textit{non-semantically} equivalent edits (as contrasted to general, semantically equivalent paraphrasing) -- this is generally beneficial: as exemplified in Fig.~\ref{fig:init_rewrite} and \ref{fig:targeted_editing}, it often enriches the prompt with concrete instructions, new concepts or descriptive details that T2I models typically respond well to. A risk, nonetheless, is that this process may also lead to edits that deviate from the user's original intent specified in $p_u$ -- for example, when a user enters a fantastical prompt, we find that through reasoning, the LLM may conclude that the requested generation is illogical or otherwise contradicts its world knowledge and proposes a ``fix'' that fundamentally differs from the user's original intent. This potential for `semantic drift' can be further compounded by the iterative editing approach that \ours takes, where modifications build upon previous ones over multiple steps.

To address this and ensure enhancements do not override the user's core requirement, we incorporate a final, dedicated ``Verify and Self-Correct'' block as a regularizer to 1) \textit{detect} whether there are core concept violations in the generated prompts from \S\ref{subsec:new_prompt} and 2) \textit{correct} violations, if any: specifically, we use the template below iteratively:

\begin{tcolorbox}[colback=mainboxbg2, colframe=mainboxborder2, coltitle=black, fonttitle=\bfseries, title=Verify \& Self-correct, boxrule=0.5mm, arc=2mm, outer arc=2mm]
\footnotesize
Your task is to first verify whether my original prompt satisfies each of the constraints specified by the user: \texttt{\{ constraints \}}

Make sure you verify \textit{each} constraint and answer "Yes" if the constraint is met or "No" otherwise, followed by a short explanation. If there is any unmet constraint, please revise the prompt to satisfy all constraints. Make sure you only modify the part of the prompt that previously led to constraint violations and bracket any revised prompt between \texttt{<answer>} and \texttt{</answer>}. If the original prompt satisfies all constraints, you may simply write "\texttt{<answer> NO\_CHANGE </answer>}".
\end{tcolorbox}
\noindent
where \texttt{\{ constraints \}} are reused DVQs from Block 2 and this refinement operation is repeated iteratively until either "\texttt{<answer> NO\_CHANGE </answer>}" is generated or a pre-defined patience (we set to 3 by default) is exhausted. With reference to Fig.~\ref{fig:mainfig}, at the end of this step, the newly generated prompt proposals (Step 10) are routed back to the T2I model and the steps repeat until convergence or any other user-defined stopping criteria (e.g., if the best prompt \& image generated so far $p^*_{:t}$ and $I^*_{:t}$ do not change after certain number of iterations).

\section{Experiments}
\label{sec:experiments}

\begin{figure}[ht!]
    \centering
    \begin{subfigure}{0.45\textwidth}
        \centering
        \includegraphics[width=\linewidth]{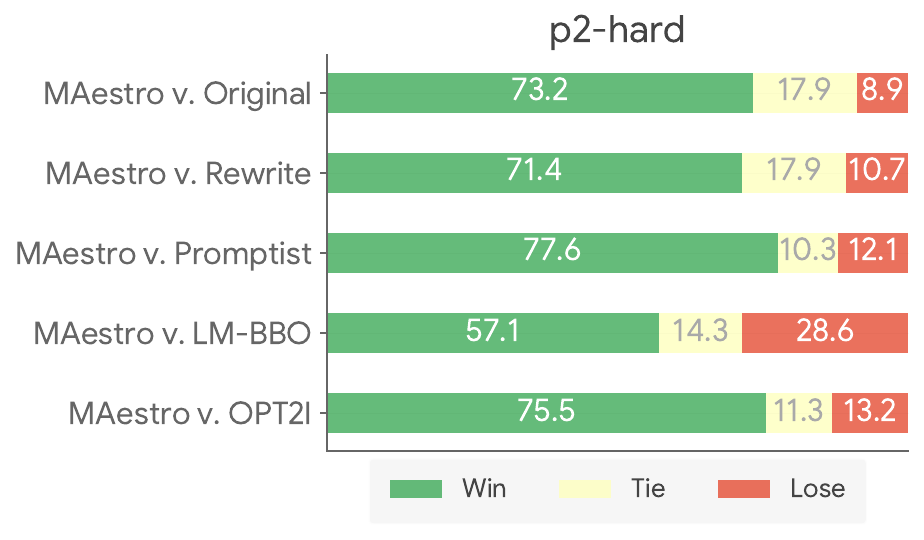}
    \end{subfigure}
    \begin{subfigure}{0.45\textwidth}
        \centering
        \includegraphics[width=\linewidth]{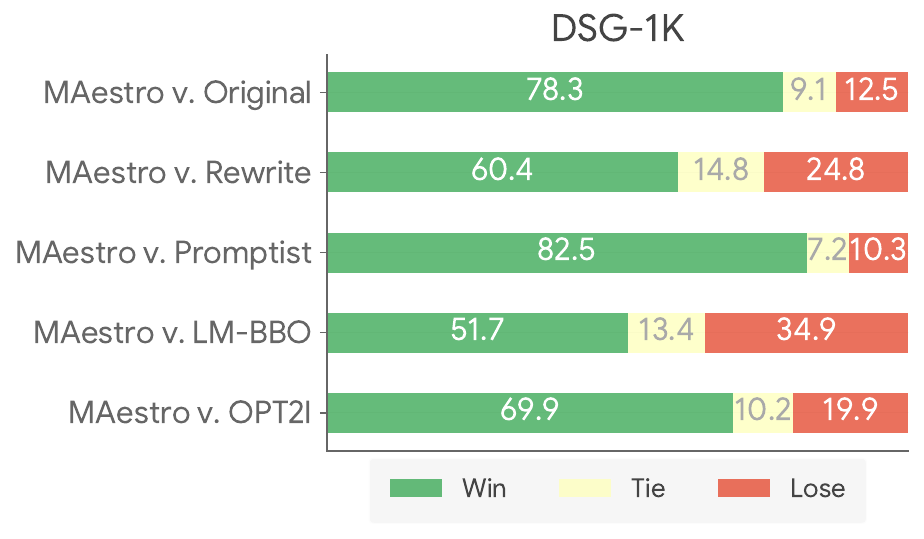}
    \end{subfigure}
    \begin{subfigure}{0.49\textwidth}
        \centering
        \includegraphics[width=\linewidth]{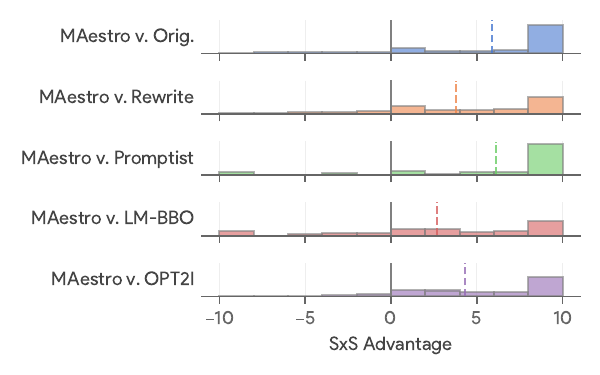}
    \end{subfigure}
    \begin{subfigure}{0.49\textwidth}
        \centering
        \includegraphics[width=\linewidth]{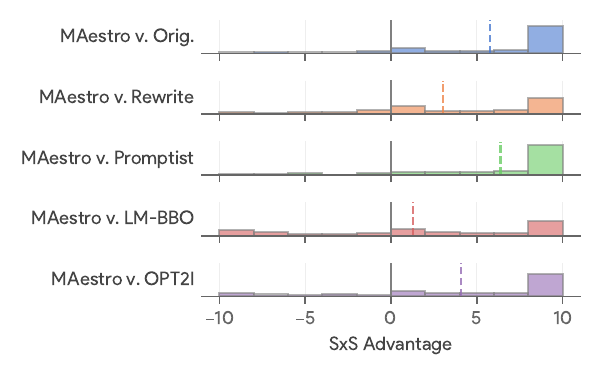}
    \end{subfigure}
    \vspace{-3mm}
    \caption{Automatic side-by-side (AutoSxS) evaluation results between \ours and the baseline methods in \textit{(left column)} p2-hard and \textit{(right column)} DSG-1K datasets using a Gemini 2.0 Flash autorater. \textit{Top row} shows the win-tie-lose breakdown and the \textit{bottom row} shows the AutoSxS advantage (i.e., \# autorater preferred \ours - \# autorater preferred the baseline) distribution: ranging from -10 (\ours is preferred in \textit{none} of the 10 trials) to 10 (\ours is preferred in \textit{all} 10 trials) and 0 denotes the autorater is overall neutral; the dashed vertical line denote the mean AutoSxS advantages across all samples.}
    \label{fig:agg_results}
\end{figure}

\rparagraph{Datasets and models.} We employed the state-of-the-art Imagen 3 T2I model~\citep{baldridge2024imagen}, available at Google Cloud. Our evaluation was conducted on two datasets:
\begin{itemize}
    \item \textbf{PartiPrompts}~\citep{yu2022scaling} comprises a curated collection of challenging user prompts. Following \citet{manas2024improving}, we focused our experiments on the samples belonging to the most challenging splits (i.e., ``fine-grained details'' and ``complex''; this subset is referred to as \textbf{p2-hard} hereafter).
    
    \item \textbf{DSG-1K}~\citep{cho2023davidsonian} is a curated dataset which itself is sampled and aggregated from several widely recognized benchmarks, including TIFA 160~\citep{hu2023tifa}, DiffusionDB~\citep{wang2022diffusiondb} / MidJourney-prompts~\citep{turc2022midjourney} (both of which were crowd-sourced and curated from real user queries), and PoseScript~\citep{delmas2022posescript}, among others. 
\end{itemize}

Recognizing the inherent strength of Imagen 3, we further implemented an additional filtering step to ensure that any observed improvements attributable different optimization strategies were not merely the result of stochastic fluctuations. For each initial prompt, with the fixed prompt, we generated images 8 times. We then used DSGScore~\citep{cho2023davidsonian} as a preliminary quality assessment tool to score each of the generated images, and we perform filtering to retain only the samples for which the model failed to achieve a perfect DSGScore of 1.0 in \textit{any} of the eight attempts. This strategy allowed us to focus on the more challenging scenarios where the model could not achieve optimal performance through repeated sampling alone. Finally, unless stated otherwise, we used Gemini 1.5~\citep{reid2024gemini} for all cases where an (M)LLM is requested (although we also experimented generation with additional models in App.~\ref{app:optimizer_model}).

\begin{table}[t!]
\centering
\caption{Mean $\pm$ standard deviation of DSGScore and rank of various methods using Gemini 2.0 Flash as the judge model. $\uparrow$: higher is better; $\downarrow$: lower is better.}
\resizebox{0.8\textwidth}{!}{
\begin{tabular}{lllll}
\toprule
Dataset   & \multicolumn{2}{c}{p2-hard}                &  \multicolumn{2}{c}{DSG-1K}          \\
Metric   & DSGScore $\uparrow$   & Rank $\downarrow$                & DSGScore $\uparrow$                 & Rank $\downarrow$                \\
\cmidrule(lr){1-1} \cmidrule(lr){2-3}  \cmidrule(lr){4-5}  
Original  & 0.826\textsubscript{$\pm$0.17}          & 4.28{$_{\pm 1.5}$}          & 0.772$_{\pm 0.18}$          & 4.36$_{\pm 1.4}$          \\
Rewrite   & 0.855\textsubscript{$\pm$0.14}          & 3.79$_{\pm 1.6}$         & 0.815$_{\pm 0.16}$           & 3.97$_{\pm 1.3}$          \\
Promptist~\citep{hao2023optimizing} & 0.873\textsubscript{$\pm$0.13}          & 3.05$_{\pm 1.7}$          & 0.849$_{\pm 0.15}$           & 2.87$_{\pm 1.8}$          \\
LM-BBO~\citep{liu2024language}    & 0.859\textsubscript{$\pm$0.15}          & 3.82$_{\pm 2.0}$          & 0.806$_{\pm 0.18}$           & 3.74$_{\pm 1.9}$          \\
OPT2I~\citep{manas2024improving}     & 0.900\textsubscript{$\pm$0.09}          & 3.41$_{\pm 1.6}$          & 0.838$_{\pm 0.15}$           & 3.25$_{\pm 1.7}$           \\
\ours (Ours)      & \textbf{0.921\textsubscript{$\pm$0.10}} & \textbf{2.65$_{\pm 1.3}$} & \textbf{0.882$_{\pm 0.13}$} & \textbf{2.83$_{\pm 1.3}$} \\
\bottomrule
\end{tabular}
}
\label{tab:dsgscore}
\end{table}
\rparagraph{Baselines.} We consider the following baselines:

\begin{itemize}
    \item \textbf{Original}: The initial, unaltered user prompt.
    \item \textbf{Rewrite}: This baseline involves using Gemini to directly rewrite the user prompt using established best practice guidelines for T2I prompting. We specifically used the guidelines provided by Google\footnote{\url{https://cloud.google.com/vertex-ai/generative-ai/docs/image/img-gen-prompt-guide}} intended to assist \textit{human} users in enhancing their prompts.
    \item \textbf{Promptist}~\citep{hao2023optimizing}: The seminal work representing train-time approaches (discussed in \S\ref{sec:analysis} to T2I prompt optimization.
    \item \textbf{Language model as a black-box optimizer (LM-BBO)}~\citep{liu2024language}, which directly uses an MLLM to iteratively improve the prompt conditioned on the last generation and the initial prompt.
    \item \textbf{OPT2I}~\citep{manas2024improving}, which uses OPRO~\citep{yang2023large}-style prompt rephrasing to optimize against a proxy, \textit{pointwise} score -- we used the DSGScore~\citep{cho2023davidsonian} as the objective in our implementation.
\end{itemize}
To ensure a fair comparison and manage computational resources, we set a maximum of 8 T2I queries for all methods that benefit from repeated sampling per generation.

\begin{wrapfigure}{r}{0.45\textwidth}
\vspace{-6mm}
\centering
\includegraphics[width=\linewidth]{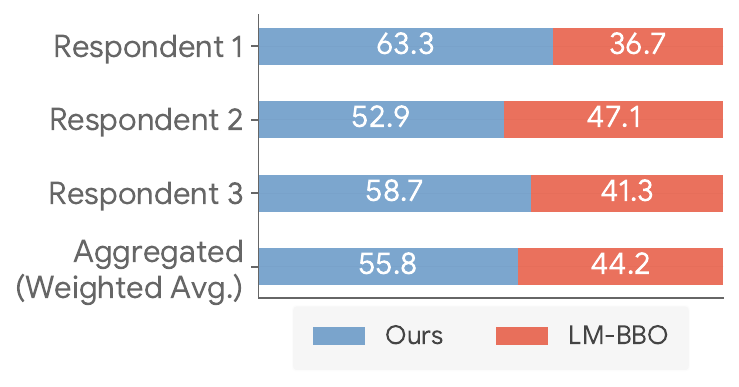}
\vspace{-7mm}
\caption{Human-preference side-by-side evaluation results between \ours and LM-BBO (next best method from automatic side-by-side evaluation in Fig.~\ref{fig:agg_results}) from 3 respondents who were given the initial user prompts, the generated image pairs, and the instruction listed in App.~\ref{app:details}. The ``\textit{Aggregated}'' column is averaged across the respondents, weighed by the number of responses submitted by each respondent.}
\label{fig:human_pref}
\end{wrapfigure}
\rparagraph{Main results.} Our evaluation incorporated both automated, model-generated metrics and real human preferences: for \textit{autorated metrics}, we used both the DSGScore as a pointwise proxy of the quality of the generated images and an automatic side-by-side (AutoSxS) pairwise comparison between generated images by two techniques conditional on the user prompts, and we present the results in Table~\ref{tab:dsgscore} and Fig.~\ref{fig:agg_results}, respectively; for both metrics, we used Gemini 2.0 Flash, which is a different and stronger judge model than the one used during the optimization (although we repeated the experiments with several other model choices to confirm the result is not unduly influenced by the judge model choice -- See App.~\ref{app:different_judge}). We also conducted a preliminary human side-by-side study where respondents were asked to choose their preferred image between \ours and LM-BBO, the next strongest method from Fig.~\ref{fig:agg_results}, and we present the results in Fig.~\ref{fig:human_pref}.

Overall, we find that \ours consistently outperforms the baselines across all metrics: for example, \ours achieves a broadly superior automatic pairwise win rate in Fig.~\ref{fig:agg_results} and a stronger DSGScore in Table~\ref{tab:dsgscore}, even outperforming OPT2I which explicitly uses the DSGScore as the optimization objective. We also find that \textbf{1)} the \textit{Rewrite} baseline, while simplistic and lightweight, proves to be remarkably effective and performs competitively compared to previously claimed state-of-the-art methods, highlighting the effectiveness of careful metaprompt design with frontier LLMs; \textbf{2)} \ours initializes with ``Rewrite'' but demonstrably surpasses it, suggesting that the additional feedback and iterative editing process provides substantial value, even with state-of-the-art generative models; and \textbf{3)} comparing Fig.~\ref{fig:agg_results} and \ref{fig:human_pref}, we observe a good agreement between automatic side-by-side comparison and human-judged comparison\footnote{Note that we did not provide a ``tie'' option in human studies -- if we were to evenly divide the ties in Fig.~\ref{fig:agg_results} to \ours and LM-BBO, the autorated preference would closely match the human preference in Fig.~\ref{fig:human_pref}.} -- this concordance further underscores the utility of the pairwise design (\S\ref{subsec:pairwise}) which aligns well with how multimodal generation quality is typically assessed in human evaluations.

\rparagraph{Qualitative analysis.} We further include selected examples of the trajectory of images generated by \ours in Table~\ref{tab:qualitative} (and additional examples in Table~\ref{app:additional_qualitative}) which showcases several interesting patterns how \ours refines image generation, often addressing nuanced aspects of the user's prompt that were not fully captured in the first attempts: for example, \ours often improves generation by addressing initial instruction following inadequacies due to \textit{underspecification} or \textit{complex or domain-specific concepts}: in Examples 3, 4 and 6 in Table~\ref{tab:qualitative} where Imagen initially generically generated a Thanos rendering, an Egyptian pyramid and a \textit{ground} telescope, \ours successfully steered the model into generating stainless steel-textured, balloon animal-like Thanos statue characteristic of \href{https://en.wikipedia.org/wiki/Jeff_Koons}{Jeff Koons}'s works (Example 3),  ``\href{https://en.wikipedia.org/wiki/Sierpi\%C5\%84ski_triangle}{Sierpiński triangle}'' pyramids with self-similar fractal features (Example 4) and \textit{space} telescope resembling Hubble with accurate text rendering on its side (zooming-in may be required to read the text) (Example 6), respectively. In other cases where the initial generation already satisfied the basic requirements in the prompt, \ours can still refine details, improve alignment with user intent, and enhance overall aesthetics as in, e.g., Example 7 (note the better overall aesthetic and emphasis on ``growth'' implied in the user prompt, as contrasted to the initial, literal interpretation of the user prompt) and Example 8 (note the more detailed glass window, better emphasis on ``calmness' and the characteristic small arms of a Tyrannosaurus rex). Overall, these qualitative examples collectively highlight our method's ability to elevate both the fidelity and the nuanced quality of generated images.

\begin{table}[t!]
\centering
\resizebox{\textwidth}{!}{
\begin{tabular}{cc}
\toprule

1. \texttt{cinematography futuristic Paris at night} & \makecell{2. \texttt{a scene with a city in the background, and a single cloud in the} \\ \texttt{foreground, with the text `contemplate the clouds' in rounded cursive}}\\
\includegraphics[width=.3\linewidth]{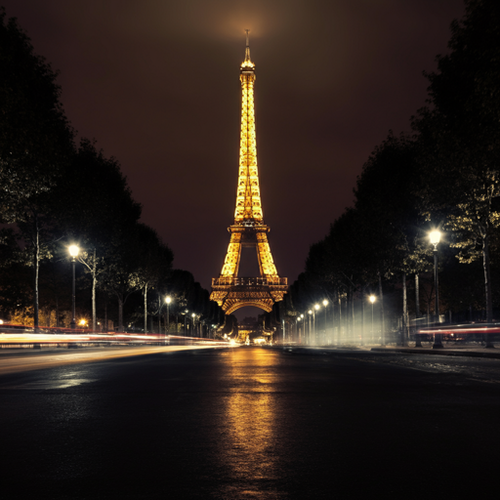} \includegraphics[width=.3\linewidth]{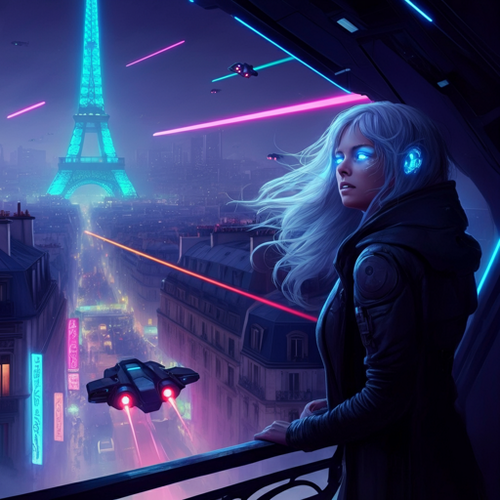} \includegraphics[width=.3\linewidth]{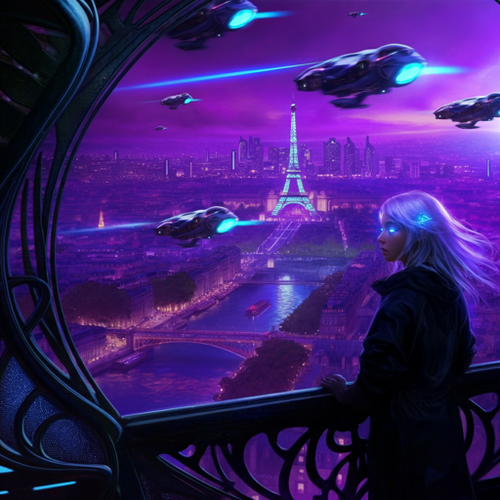} & 
\includegraphics[width=.3\linewidth]{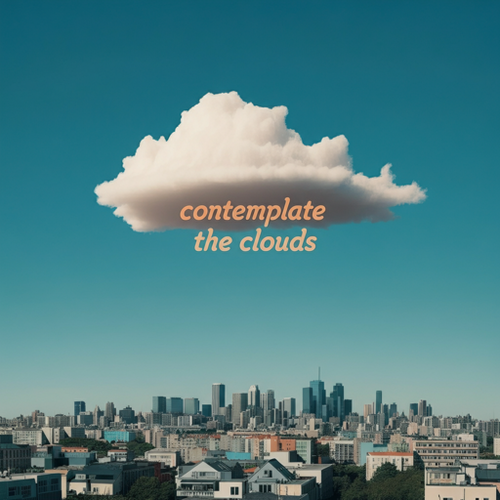} \includegraphics[width=.3\linewidth]{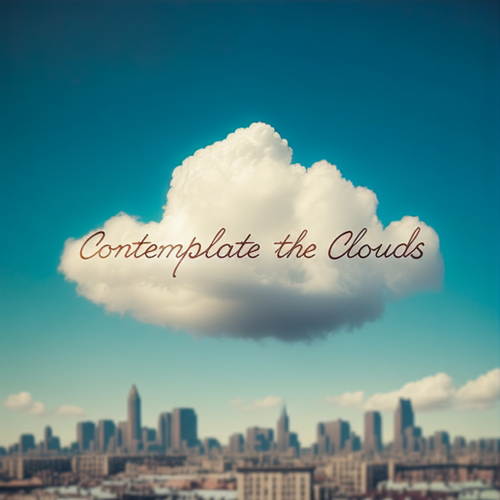} \includegraphics[width=.3\linewidth]{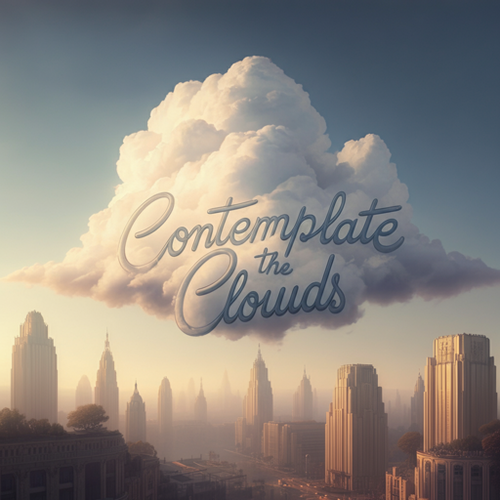}\\ 
3. \texttt{statue of Thanos made by Jeff Koons} & 4. \texttt{Sierpiński triangle pyramids of Egypt}\\
\includegraphics[width=.3\linewidth]{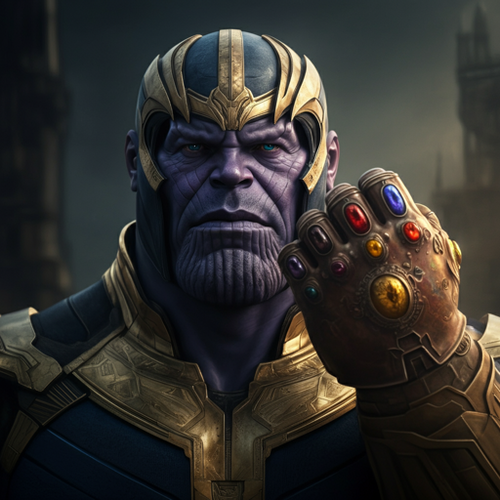} \includegraphics[width=.3\linewidth]{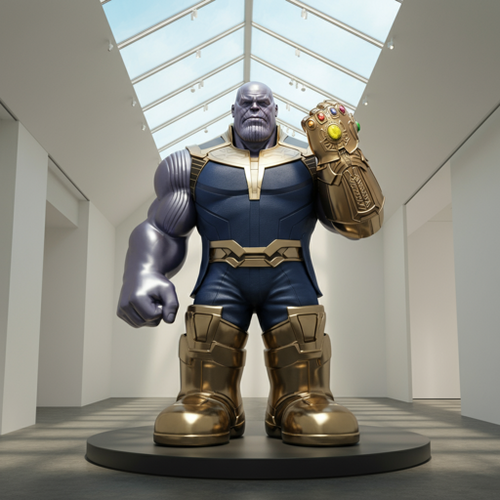} \includegraphics[width=.3\linewidth]{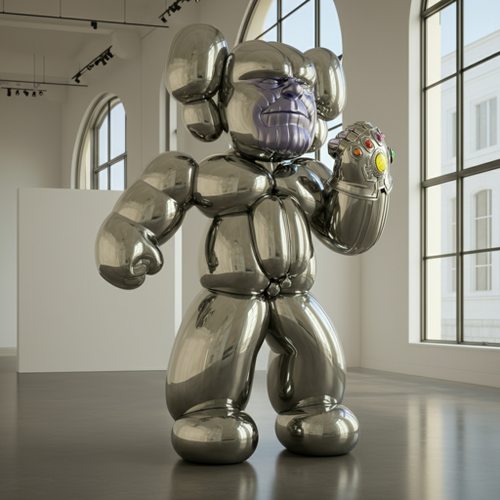} & 
\includegraphics[width=.3\linewidth]{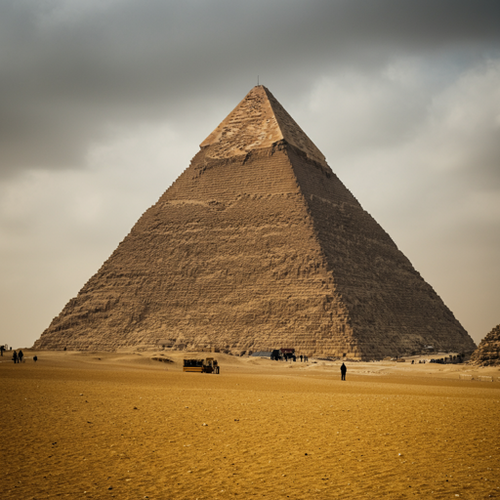} \includegraphics[width=.3\linewidth]{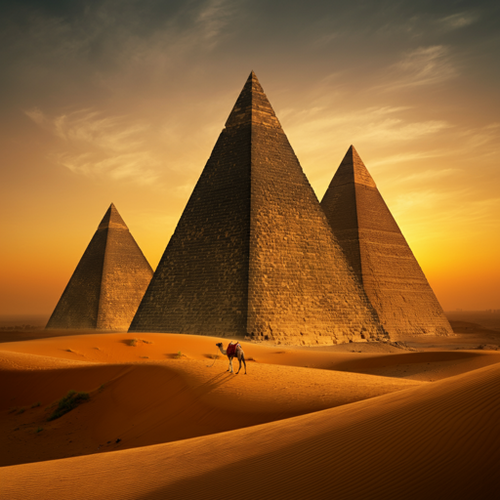} \includegraphics[width=.3\linewidth]{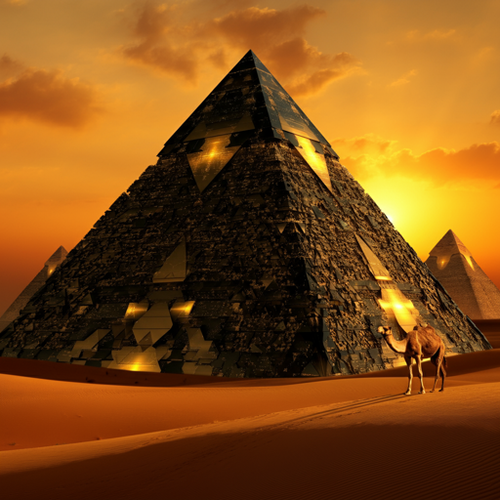}
\\
\makecell{5. \texttt{the view from one end of a bench in a park, looking}\\\texttt{at the sky with the text `imagine the outcome' in the sky}} & \makecell{6. \texttt{the hubble telescope and the milky way, with the text} \\ \texttt{`the universe is a mystery, but we are here to solve it'}}\\
\includegraphics[width=.3\linewidth]{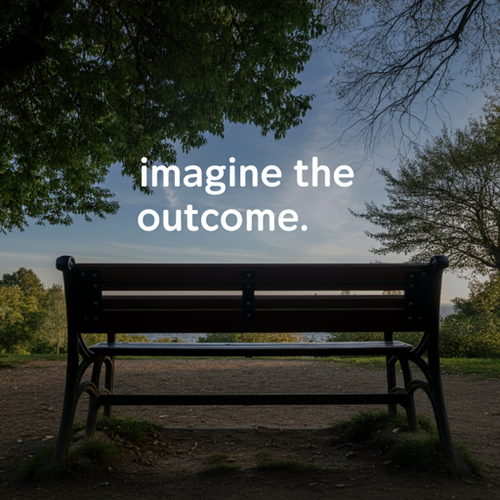} \includegraphics[width=.3\linewidth]{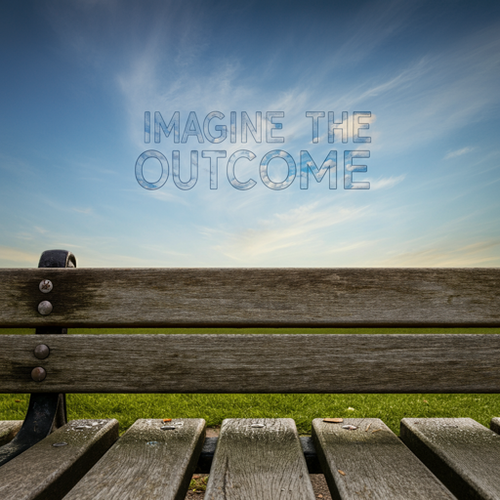} \includegraphics[width=.3\linewidth]{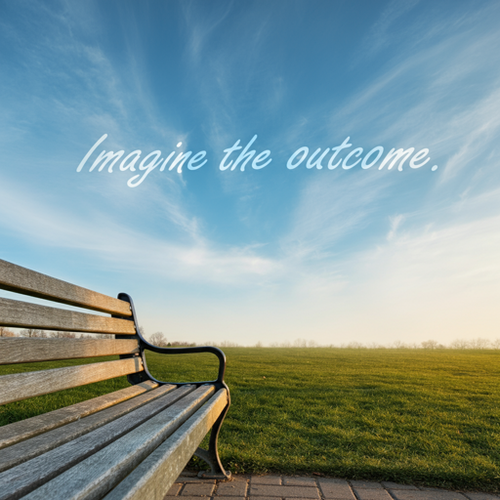} & 
\includegraphics[width=.3\linewidth]{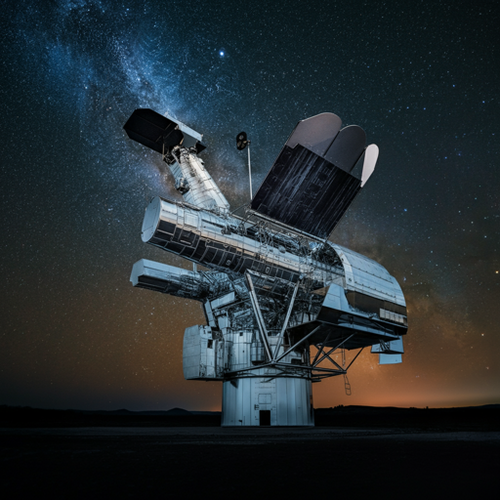} \includegraphics[width=.3\linewidth]{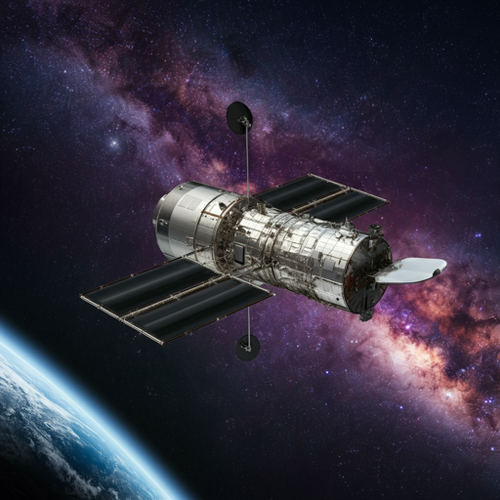} \includegraphics[width=.3\linewidth]{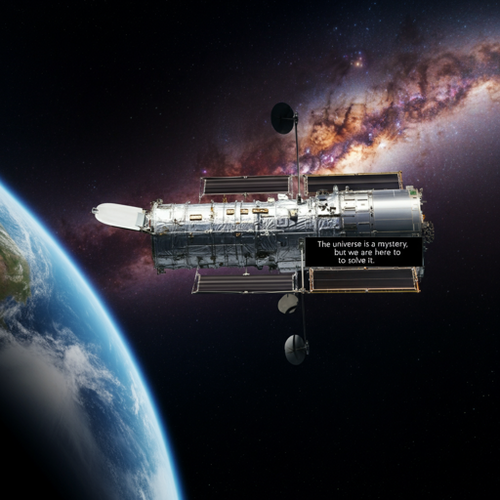}
\\ 
\makecell{7. \texttt{a picture of multiple trees at various stages of development,} \\ \texttt{with the caption `growth is a continuous process'}} & 8. \texttt{a stained glass window depicting a calm tyrannosaurus rex}\\

\includegraphics[width=.3\linewidth]{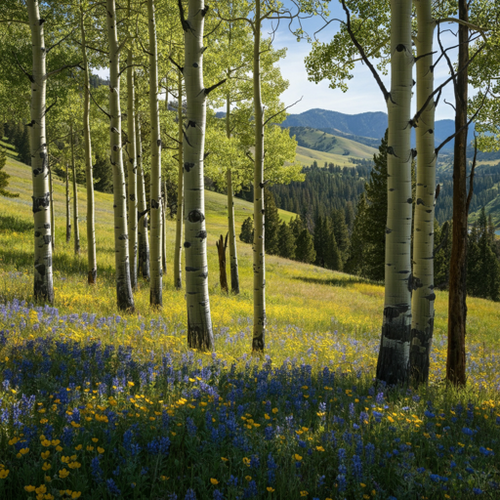} \includegraphics[width=.3\linewidth]{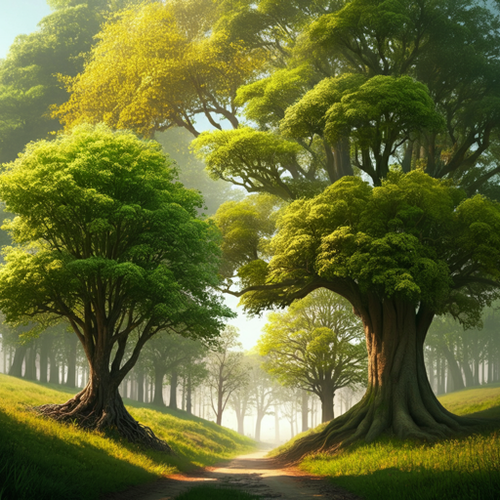} \includegraphics[width=.3\linewidth]{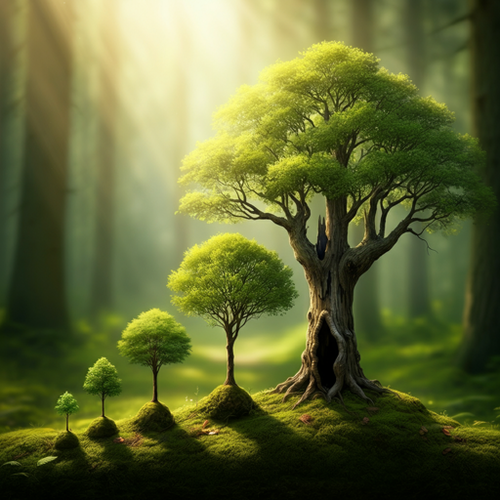} &
\includegraphics[width=.3\linewidth]{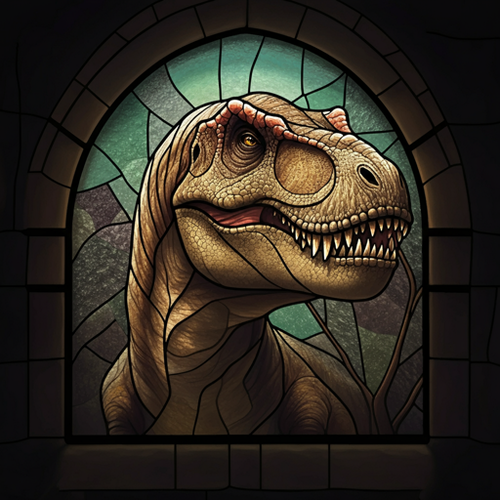} \includegraphics[width=.3\linewidth]{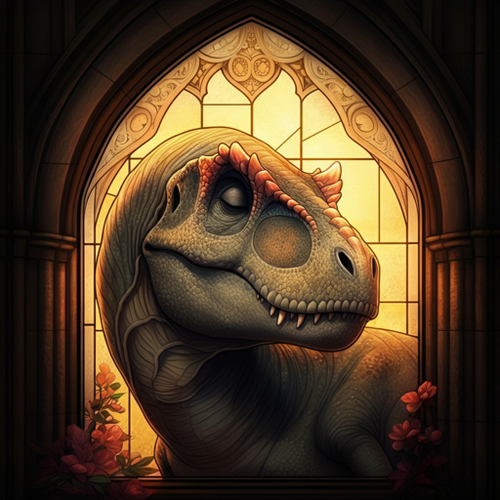} \includegraphics[width=.3\linewidth]{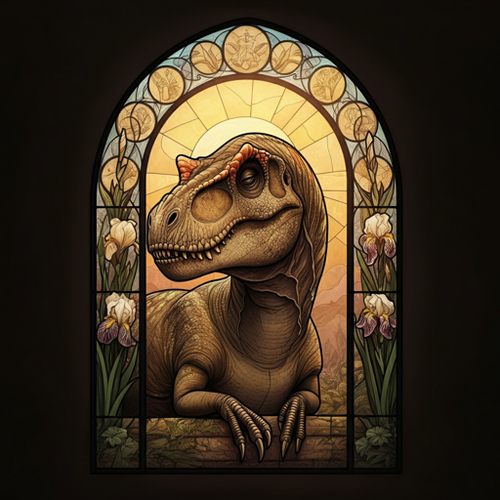}
\\
\bottomrule
\end{tabular}
\caption{Qualitative examples of the image trajectories generated by \ours using different \textit{original} user prompts (which are given at the top of each image row). Each row, from left to right, shows 1) image generated using the original, unaltered user prompt; 2) an intermediate generation, and 3) the final, optimized image returned by \ours after to up to 8 iterations. Refer to App.~\ref{app:additional_qualitative} for additional visualizations and the text prompts that generated the presented images.}
\label{tab:qualitative}
}
\vspace{-2mm}

\label{fig:qualitative_study}

\end{table}

\begin{figure}[t!]
    \centering
    \begin{subfigure}{0.49\textwidth}
        \centering
        \includegraphics[width=\linewidth]{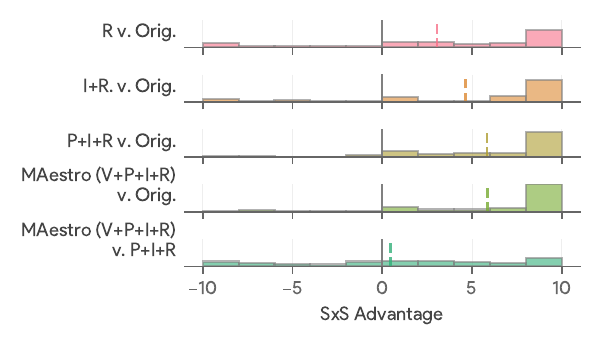}
    \end{subfigure}
    \begin{subfigure}{0.49\textwidth}
        \centering
        \includegraphics[width=\linewidth]{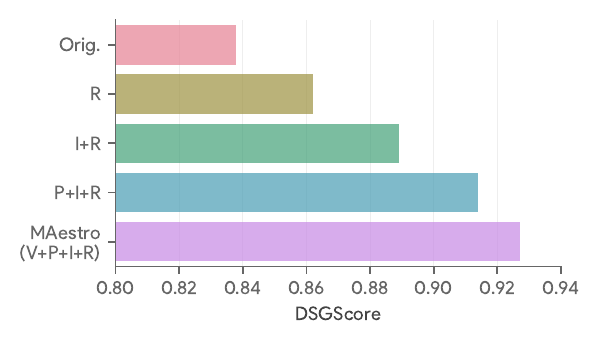}
    \end{subfigure}
    \vspace{-3mm}
    \caption{Ablation studies showing \textit{(left)} pairwise AutoSxS advantage distributions (note that first 4 rows are compared w.r.t. the images generated by the original prompt; the final row compares \ours (full algorithm) against P+I+R variant; \textit{(right)} pointwise DSGScore of the different algorithmic variants.}
    \label{fig:ablation}
\end{figure}

\rparagraph{Ablation studies.} To understand the contribution of different components of our system, we also conducted ablation studies on a dataset randomly subsampled from p2-hard and DSG-1K and we study the following \ours variants in Fig.~\ref{fig:ablation}:
\begin{enumerate}
    \item \texttt{R}: Prompts generated by the initial \textbf{R}ewriting (\S\ref{subsec:init}) only;
    \item \texttt{I+R}: Prompts generated by initial \textbf{R}ewriting and the \textbf{I}terative improvements (\S\ref{subsec:init} \& \S\ref{subsec:new_prompt}) where we always iterate upon the \textit{last generation} (i.e., without tracking \textit{the best generation so far} discussed in \S\ref{subsec:pairwise} or the self-verification and correction mechanism in \S\ref{subsec:intent_grounding});
    \item \texttt{P+I+R}: Same as above, but with \textbf{P}airwise comparator (\S\ref{subsec:pairwise}) added to track and improve upon the best generation so far (rather than simply iterating on the last iteration);
    \item \texttt{\ours (V+P+I+R)}: the full \ours algorithm; the difference w.r.t. the previous method is the introduction of the self-\textbf{V}erification component (\S\ref{subsec:intent_grounding}).
\end{enumerate}

By default, we compared each of the variants against the image generated with the original user prompts (\texttt{Orig.}), although we added one experiment directly comparing \texttt{\ours (V+P+I+R)} and \texttt{P+I+R} given that they achieved similar advantage against \texttt{Orig.} (5.88 vs 5.84); we find that the full \ours still has a modest advantage over \texttt{P+I+R} when compared head-to-head and a bigger improvement in terms of DSGScore--this is unsurprising as the goal of self-verification component is to ground the newly generated prompts to the user intent, which is reflected by the improvement in DSGScore that tracks the prompt-image consistency. On the other hand, the addition of the other components unequivocally improved both AutoSxS score and the DSGScore, demonstrating the utility of each component of \ours in improving the overall performance.

\section{Conclusion}
\label{sec:conclusion}

We present \ours, a novel multi-agent system that empowers T2I models to autonomously refine their output through iterative prompt adjustments. By leveraging multi-agent critique with interpretable feedback and a pairwise MLLM-as-a-judge objective for robust evaluation, \ours significantly improves image quality over initial prompts and state-of-the-art techniques, with benefits scaling with MLLM advancements. We believe the scope for future work is ample: for example, while our current evaluations focus on the Imagen model, the framework's model-agnostic design invites future testing across diverse T2I systems which is an immediate next step. Second, while we have focused on T2I generation, the core principles of \ours also show promise for general generative tasks lacking ground truths and that might benefit from targeted supervision for editing, including but not limited to, for example, \textit{audio} or \textit{video} synthesis. Lastly, while \ours is currently a test-time algorithm, the generated improvement trajectories could be integrated into the T2I model training pipeline, enabling autonomous self-improvement beyond prompt refinement -- we leave these important explorations to future work.

\bibliographystyle{abbrvnat}
\nobibliography*
\bibliography{iclr2025_conference}
\newpage
\appendix
\section{Implementation Details}
\label{app:details}

In this section, we describe the implementation details, including the prompt templates used in different stages of \ours aside from the ones listed in the main text and the specific hyperparameters used in both \ours and the baselines. 

\rparagraph{Initialization (\S\ref{subsec:init})}
We show the prompt template for the initial LLM rewrite below, using the Jinja2 syntax with the brace-enclosed variables to be replaced with actual values during compilation, where \texttt{current\_prompt} denotes the original, user prompt $p_u$ with \texttt{n\_prompt} set to 1 by default. The output is extracted by the regex \texttt{<PROMPT>(.*?)</PROMPT>}. Note that we also use the prompt below for the `Rewrite' baseline.

\begin{tcolorbox}[colback=yellow!15!white, colframe=yellow!50!black, coltitle=black, fonttitle=\bfseries, title=Prompt template for initial rewrite (\S\ref{subsec:init}), boxrule=0.5mm, arc=2mm, outer arc=2mm]
\scriptsize
Hi Gemini, I want to create an image by giving a prompt to Imagen, a text-to-image model.

Here is my current prompt:
"\{ \texttt{current\_prompt} \}"

Please generate an improved prompt based on the following guidelines for Imagen:

1. Make sure the prompt contains the elements of \textbf{subject}, \textbf{context}, and \textbf{style}. Specifically, \\
  - Subject: The first thing to think about with any prompt is the subject: the object, person, animal, or scenery you want an image of. \\
  - Context and background: Just as important is the background or context in which the subject will be placed. Try placing your subject in a variety of backgrounds. For example, a studio with a white background, outdoors, or indoor environments. \\
  - Style: Finally, add the style of image you want. Styles can be general (painting, photograph, sketches) or very specific (pastel painting, charcoal drawing, isometric 3D).

2. Use descriptive language: Employ detailed adjectives and adverbs to paint a clear picture for Imagen 3.

3. Provide context: If necessary, include background information to aid the AI's understanding.

4. Reference specific artists or styles: If you have a particular aesthetic in mind, referencing specific artists or art movements can be helpful.

5. Use prompt engineering tools: Consider exploring prompt engineering tools or resources to help you refine your prompts and achieve optimal results.

6. Enhancing the facial details in your personal and group images: \\
  - Specify facial details as a focus of the photo (for example, use the word "portrait" in the prompt).

Please now generate \{ \texttt{n\_prompt} \} improved prompt(s) and enclose any new prompt between \texttt{<PROMPT>} and \texttt{</PROMPT>}.
If you believe the current prompt is already satisfactory, simply respond with \texttt{"<PROMPT> NO\_CHANGE </PROMPT>"}.
\end{tcolorbox}

For the DVQs, we re-used the instructions and few-shot examples provided by \citet{cho2023davidsonian} to decompose a given user prompt to a list of DVQs. These DVQs were generated once, and are subsequently fixed for all methods that use these questions (e.g., \ours and OPT2I).

\rparagraph{Pairwise comparison (\S\ref{subsec:pairwise})} While we have already included the prompt for pairwise comparison in \S\ref{subsec:pairwise} in the main text, here we give additional implementation details: as discussed, at the end of each prompt generation step, we perform binary tournament amongst the newly generated prompt-image pairs first (if more than one image is generated in the step): we ask the MLLM to choose a preferred image between a pair of images (denoted as Image A and Image B) for $2n$ times (we set $n=3$ by default) in parallel (for the first $n$ times, the model will place Image A first and for the remaining $n$ times, the model will place Image B first to eliminate position bias). This process continues until there is a single candidate prompt remaining, which is denoted the \textit{best prompt so far}.

\rparagraph{Prompt generation (\S\ref{subsec:new_prompt})}
Below we show the prompt templates given to the LLM to generate new prompt(s) based on previous feedback (\S\ref{subsec:new_prompt}), and we use the following templates for \textit{targeted editing}, which consists of 1) visual question answering by an MLLM, given the DVQs in the previous step and 2) generating new prompts given by the DVQs along with the LLM's answers. For the first step, by default we break it into two LLM calls: firstly asking the model to give an answer \textit{without explaining}, followed by another call instructing the model to \textit{rationalize} its answer in a posthoc manner -- it is worth noting that a (potentially better) alternative is to simply use chain-of-thought (CoT) style prompting and retain the rationale part, but we opted for posthoc rationalization to ensure fairness of comparison against the baselines (which relies on the final answer given by the LLM only) and to avoid the discrepancy introduced by using the CoT-style template, which may influence the MLLM prediction.

\begin{tcolorbox}[colback=blue!15!white, colframe=blue!50!white, coltitle=black, fonttitle=\bfseries, title=Prompt template for DVQ \textit{question-answering} (\S\ref{subsec:new_prompt}), boxrule=0.5mm, arc=2mm, outer arc=2mm]
\scriptsize
Answer only with `yes' or `no'. Do not give other outputs or punctuation marks. \\
Image: \{ \texttt{image} \} \\
Question: \{ \texttt{dvq[i]} \}
\end{tcolorbox}

\begin{tcolorbox}[colback=blue!15!white, colframe=blue!50!white, coltitle=black, fonttitle=\bfseries, title=Prompt template for DVQ \textit{rationalization} (\S\ref{subsec:new_prompt}), boxrule=0.5mm, arc=2mm, outer arc=2mm]
\scriptsize
You are an expert in examining images generated by text-to-image generation models. The user specifies a list of questions, each of which contains a property or characteristic that the generated image should have and the goal is to make sure that when an reviewer reviews the generated image, they will answer ``Yes" to each of these questions. However, the reviewer answered ``No" to the question below to the current image: \\
Image: \{ \texttt{image} \} \\
Question: \{ \texttt{dvq[i]} \}
\end{tcolorbox}
\noindent
For the actual prompt generation step, we used the following template for the \textit{targeted editing} generator, where \texttt{best\_prompt\_so\_far} is the best prompt so far $p^*_{:t}$ as determined by the Comparator (\S\ref{subsec:pairwise}); \texttt{n\_questions} denotes the number of decomposed visual questions (DVQs) (discussed in \S\ref{subsec:init}); \texttt{dvq} is a list of actual DVQ, and \texttt{feedback} the MLLM response to the rationalization question above:

\begin{tcolorbox}[colback=green!15!white, colframe=green!70!black, coltitle=black, fonttitle=\bfseries, title=Prompt template for targeted editing (\S\ref{subsec:new_prompt}), boxrule=0.5mm, arc=2mm, outer arc=2mm]
\scriptsize
You are an expert in prompt engineering for text-to-image generation models, which take a text prompt as input and generate images depicting the prompt as output. Your task is to improve upon my current prompt based on feedback I obtained from human reviewers.

The user has specified a list of questions, each of which contains a property or characteristic that the generated image should have and the goal is to make sure that when a reviewer reviews the generatd image, they will answer `Yes' to all the questions. However, they have answered `No' to one or more questions and provided some feedback, and I'd like you to help me improve my existing prompt, taking into account their feedback.

My current prompt is:\\
`\{ \texttt{best\_prompt\_so\_far} \}'

But a reviewer looked at my image and answered `No' to the following question(s). Below are the questions and their feedback, which contains explanations why they believed the image does not satisfy the requirement in the question(s) and their suggestions for improvements:

\{\% \texttt{for i in range(n\_questions)} \%\} \\
Question \{\texttt{ i + 1 }\}: \\
\{ \texttt{dvq[i]} \} \\
Feedback \{ \texttt{i + 1} \}: \\
\{ \texttt{feedback[i]} \} \\
\{ \texttt{end\_for} \} \\

Based on your analysis and the reviewer's opinion, now improve my prompt by incorporating all their feedback points into {{ \texttt{new\_solutions} }} new prompt(s) so that the reviewer will now answer `Yes' to the question(s) when I generate a new image with the new prompt(s). Pay close attention to any points that they mentioned are lacking in the current image, and ensure you incorporate all these information in your revised prompts. Make sure the revised prompts are concise and effective. You may be creative in composing the new prompts, adding new elements, emphasize existing elements, or shuffle the order of the sentences to \textit{emphasize the previously missing characteristics at the beginning of the prompt}.
Finally, make sure each prompt you generated is bracketed between \texttt{<PROMPT>} and \texttt{</PROMPT>}
\end{tcolorbox}

For \textit{implicit improvement} generator, we use an adapted version of the prompt from \citet{liu2024language} where we 1) make sure it is always iterated on the \textit{best image so far} (as opposed to the \textit{last generation}) and 2) we added an option for the model to early terminate with \texttt{<PROMPT> NO\_CHANGE </PROMPT>} if it decides that the current generation is already satisfactory (the second change is also applied to our implementation of the LM-BBO baseline, since we believe it leads to a better performance than asking the model to keep editing without any termination criterion). 

\begin{tcolorbox}[colback=orange!15!white, colframe=orange!70!black, coltitle=black, fonttitle=\bfseries, title=Prompt template for implicit improvement (\S\ref{subsec:new_prompt}), boxrule=0.5mm, arc=2mm, outer arc=2mm]
\scriptsize
Hi Gemini, I want to create an image featuring the topic of `\{ \texttt{user\_prompt} \}'

Here is my current prompt: \\
`\{ \texttt{best\_prompt\_so\_far} \}'

I asked Imagen to generate an image using the above prompt. Does it correctly reflect the topic? 

Image: \\
\{ \texttt{best\_image\_so\_far} \}

If yes, please simply respond with "\texttt{<PROMPT> NO\_CHANGE </PROMPT>}".
If no, please help me modify the prompt by generating \{ \texttt{n\_prompts} \} new prompt(s) and respond with any new prompt in between \texttt{<PROMPT>} and \texttt{</PROMPT>}.
\end{tcolorbox}

For all experiments using \ours, we always set \texttt{n\_prompts} to 1 for both templates and at each generation we invoke both targeted editing and implicit improvement once in parallel -- this thus gives two proposed prompts per each iteration. We set the maximum number of T2I calls to be 8 (which translates to 4 total steps per query). For the LM-BBO baseline, we only use the implicit improvement template \textit{upon the last generation} with \texttt{n\_prompts} set to 1 until the maximum T2I model call (8) is reached or the early termination is triggered (i.e., model generating \texttt{<PROMPT> NO\_CHANGE </PROMPT>}).

\rparagraph{Human preference studies.} For human preference study, we collected responses from 3 human respondents (who are all authors' colleagues) where we show the human respondents the initial, user prompt and a pair of images (but not the optimized prompts) generated from LM-BBO (the next best method from the automated side-by-side studies) and \ours in an interative Google Colab; the ordering of the images is randomized per pair to ensure that the respondents are unable to guess the method that generated the images. We provided the following instructions to the respondents:

\begin{tcolorbox}[colback=red!15!white, colframe=red!70!white, coltitle=black, fonttitle=\bfseries, title=Instructions for human preference study, boxrule=0.5mm, arc=2mm, outer arc=2mm]
\footnotesize
You will be shown \textbf{a pair of images} (i.e., Left Image and Right Image) and an user prompt (in most cases, the prompt you see is an actual user prompt that was entered into a T2I system (e.g., Midjourney or Imagen)).
\\
Please click "Left Image" or "Right Image" as your preferred image \textit{given the user prompt} -- the ordering ("Left" or "Right") is randomized between \texttt{REFERENCE\_DIR} and \texttt{CANDIDATE\_DIR} that you selected above.
\\
Try to complete as many selections as possible, but you can stop at any time as your response will be immediately stored.
\\
Some prompts specify command line arguments such as aspect ratio or resolution -- please ignore these options when judging the images the model used to generate the provided images do not account for these options.
\\
In some cases, the user prompt may contain niche / domain information (e.g., specification of artistic styles or they'd like the image to imitate generations from certain engines). In these cases, please take your time and do research if needed, as your previous response was already saved.
\\
\textbf{Please always select an image even if there is no clear winner between the two options} -- we deliberately did not provide a tie option. You may consider factors such as overall coherence, high-level aesthetic, and how close the generate images follow the original user prompt to choose a better image. \textbf{If the two images are still similar even after considering the above or, in some cases, the two provided images are identical, please simply pick an image randomly}.
\end{tcolorbox}
where \texttt{REFERENCE\_DIR} and \texttt{CANDIDATE\_DIR} by default point to the directory saving the images generated from LM-BBO and \ours, respectively.

\section{Additional Experiments}
\label{app:additional_experiments}

\subsection{Different Judge Models}
\label{app:different_judge}

\begin{figure}[ht!]
    \centering
    \begin{subfigure}{0.32\textwidth}
        \centering
        \includegraphics[width=\linewidth]{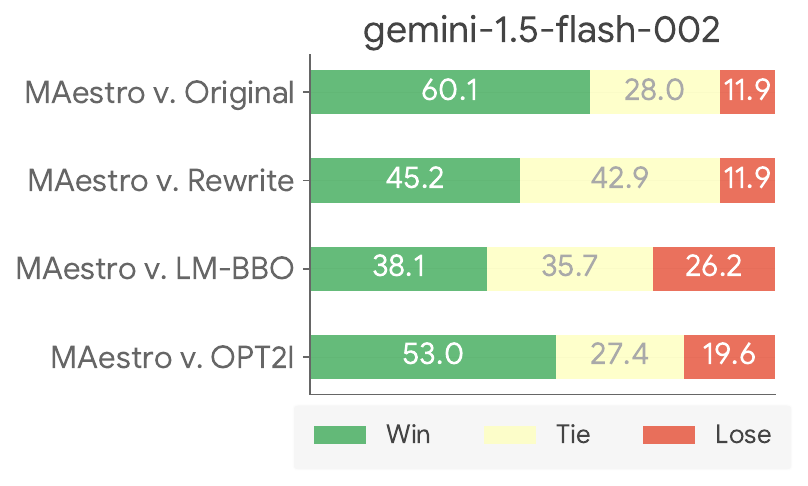}
    \end{subfigure}
    \begin{subfigure}{0.32\textwidth}
        \centering
        \includegraphics[width=\linewidth]{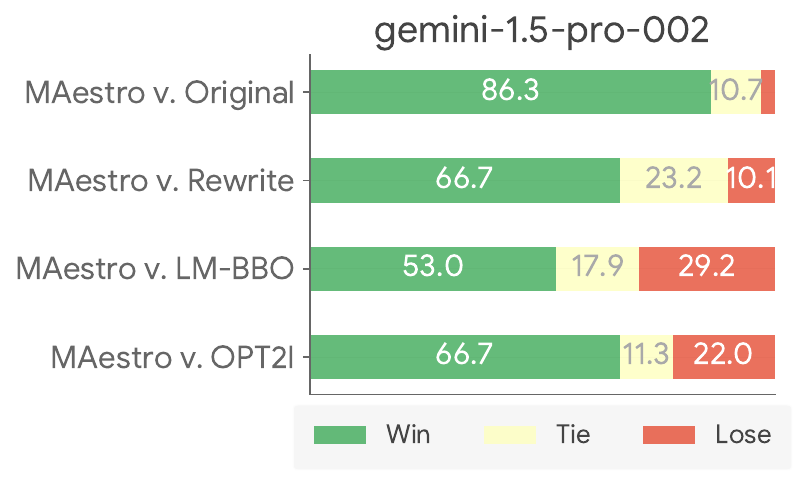}
    \end{subfigure}
    \begin{subfigure}{0.32\textwidth}
        \centering
        \includegraphics[width=\linewidth]{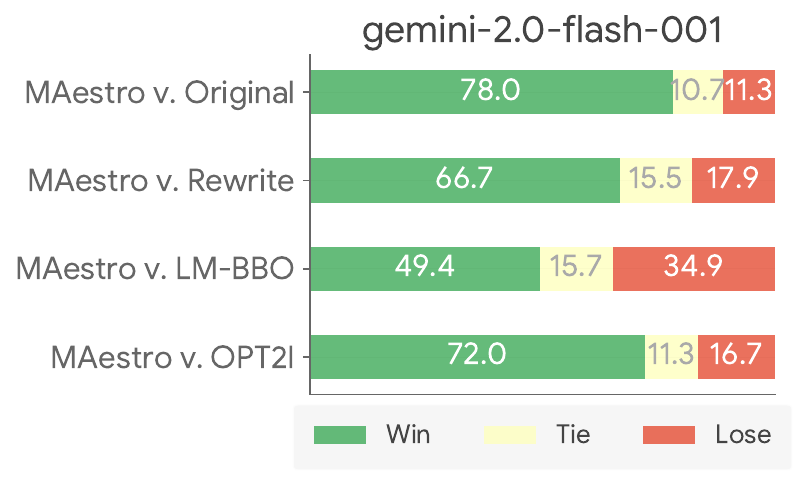}
    \end{subfigure}
    \caption{AutoSxS win-tie-lose breakdown using different \textit{judge} models on a dataset subsampled from p2-hard and DSG-1K.}
    \label{fig:different_judges}
\end{figure}

In this section, we explore the use of different \textit{judge} models and their influence on the final results and we show the results in Fig.~\ref{fig:different_judges} where we use Gemini 1.5 Pro, Gemini 1.5 Flash and Gemini 2.0 Flash as pairwise judge in the AutoSxS studies. Overall, we find that the relative strength of the methods as implied by the side-by-side comparison is consistent, although Gemini 1.5 Flash noticeably generates a larger fraction of ``tie'' ratings, likely due to its weaker multimodal understanding capabilities and stronger positional bias (e.g., a tie will be returned if a judge model always prefers the first image or second image) compared to the other judge models.

\subsection{Different Optimizer Models}
\label{app:optimizer_model}

\begin{figure}[th!]
\centering
    \includegraphics[width=.5\linewidth]{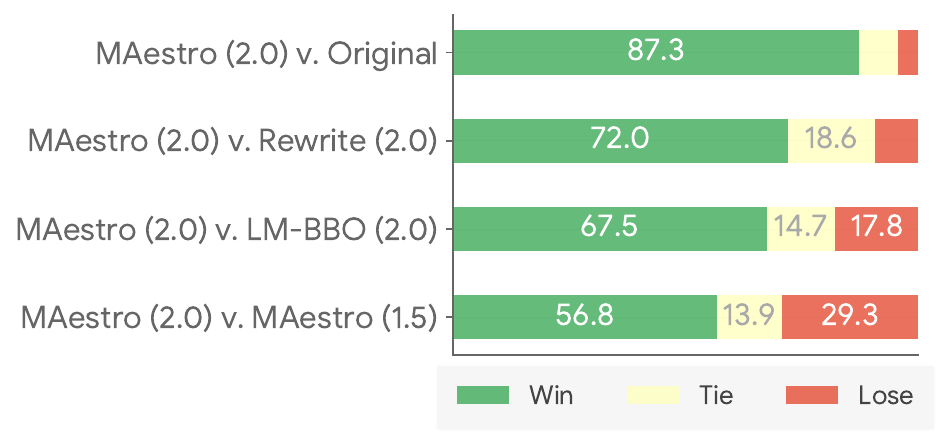}
    \caption{AutoSxS win-tie-lose breakdown using different \textit{optimizer} models (methods using Gemini 2.0 and 1.5 are marked with 2.0 and 1.5, respectively) on DSG-1K.}
    \label{fig:different_optimizers}
\end{figure}
We also explore using different \textit{optimizer} model to understand how the models with different multimodal understanding caliber would affect the overall effectiveness of the \ours pipeline. In Fig.~\ref{fig:different_optimizers} we show the results of \ours (and baselines) using Gemini 2.0 Flash optimizer model as opposed to Gemini 1.5 optimizer used by default in the main text. Overall, we find that \ours improves with better foundation model as the optimizer model, with the win rate improving both over the baseline methods using the corresponding optimizer model and \ours with a weaker model (the last row of Fig.~\ref{fig:different_optimizers}). 

\subsection{Additional Qualitative Results}
\label{app:additional_qualitative}

In this section, we present additional qualitative results (Table~\ref{tab:additional_qualitative}) to complement Table ~\ref{tab:qualitative} in the main text. We also show the optimized text prompts that led to the generated images in Tables~\ref{tab:qualitative} and \ref{tab:additional_qualitative} in Table~\ref{}.

\begin{table}[t!]
\centering
\resizebox{\textwidth}{!}{
\begin{tabular}{cc}
\toprule

\makecell{1. \texttt{realistic detailed 1950 s style movie poster}\\ \texttt{retrofuturistic giving a presentation to a room of } \\ \texttt{scientists. moebius, brom, ian miller, moody vibrant colors}} & \makecell{2. \texttt{Superman with a spiderman mask}}\\
\includegraphics[width=.3\linewidth]{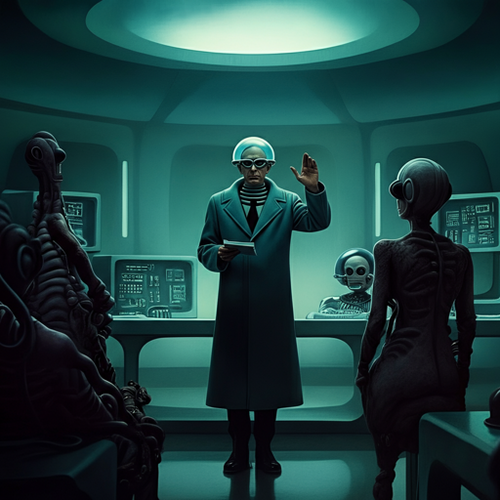} \includegraphics[width=.3\linewidth]{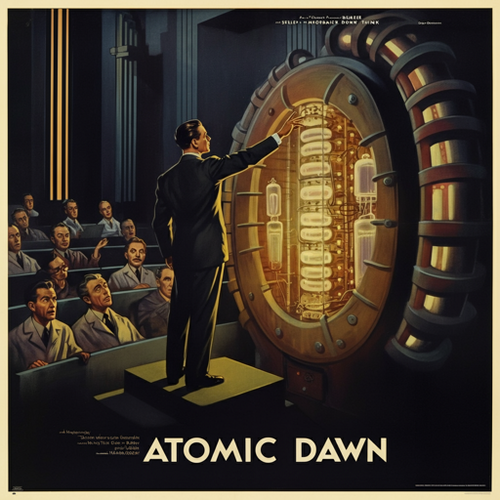} \includegraphics[width=.3\linewidth]{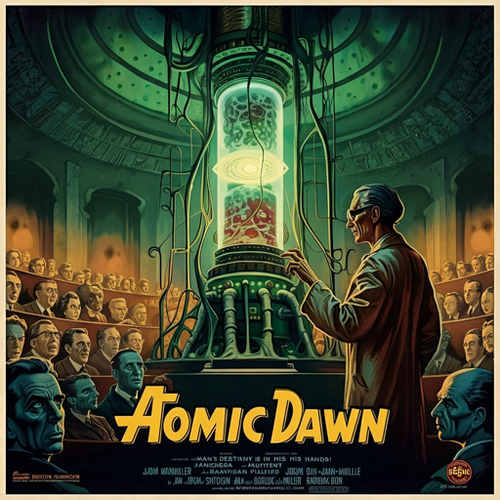} & 
\includegraphics[width=.3\linewidth]{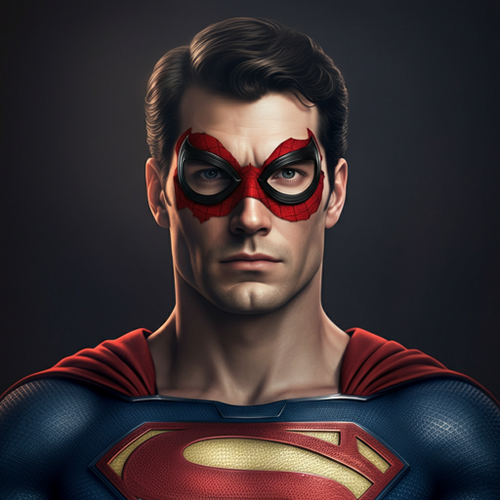} \includegraphics[width=.3\linewidth]{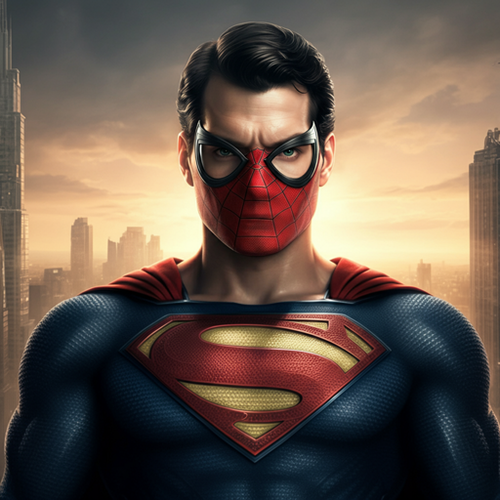} \includegraphics[width=.3\linewidth]{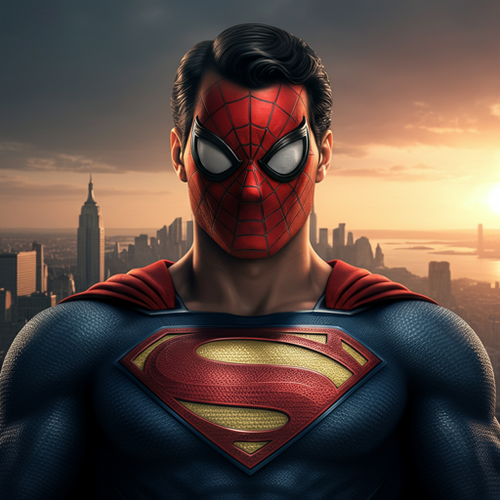}\\ 
\makecell{3. \texttt{man made of swirling smoke and wind - blown snow, holding a portal to} \\ \texttt{the universe, photorealistic, highly detailed, octane render}} & \makecell{4. \texttt{anthropomorphic furry wolf in armor}  \\  \texttt{fighting in a battlefield, 1900s picture}}\\
\includegraphics[width=.3\linewidth]{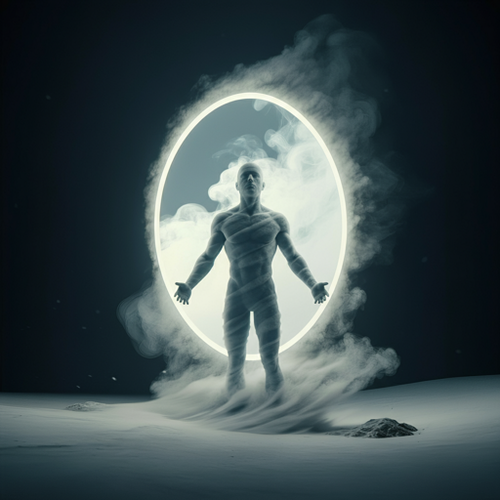} \includegraphics[width=.3\linewidth]{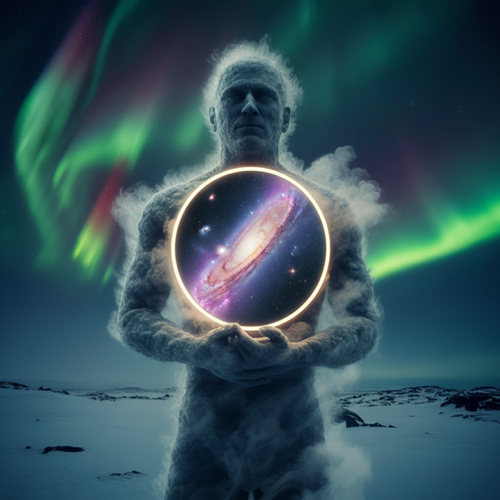} \includegraphics[width=.3\linewidth]{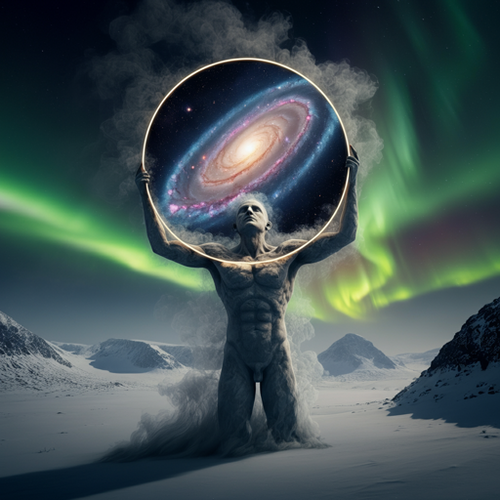} & 
\includegraphics[width=.3\linewidth]{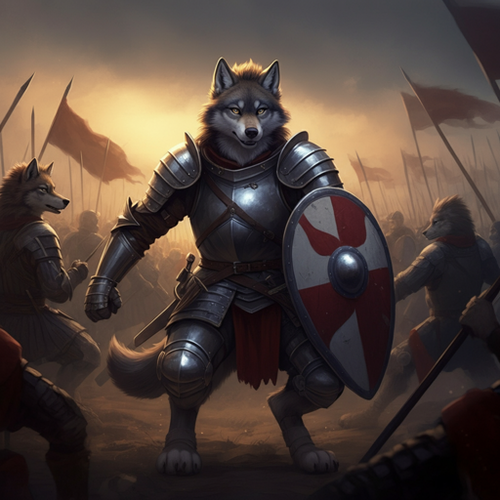} \includegraphics[width=.3\linewidth]{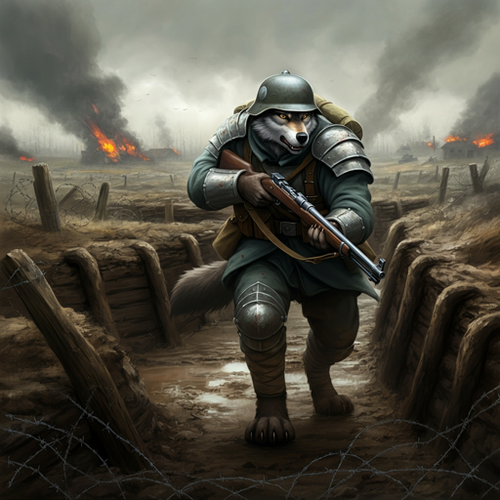} \includegraphics[width=.3\linewidth]{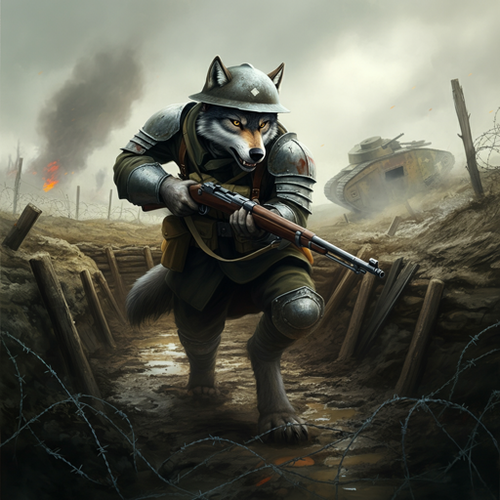}
\\
\makecell{5. \texttt{set of nine abstract bicycles Ilustrace}} & \makecell{6. \texttt{the words 'Art is never finished, only continued' in paint splatters} \\ \texttt{ on a white background graffiti art, edge of nothingness} \\  \texttt{love, muddy colors, colored woodcut, beautiful, spectral color"}}\\
\includegraphics[width=.3\linewidth]{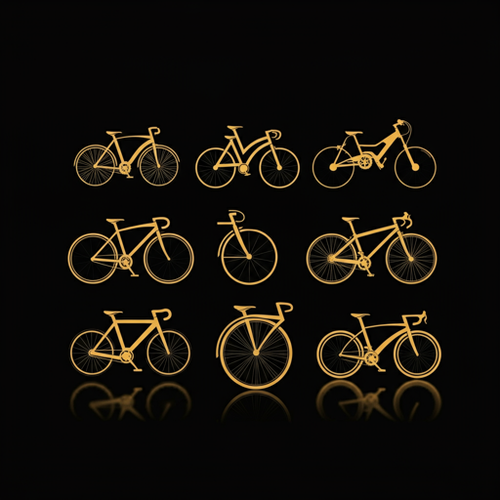} \includegraphics[width=.3\linewidth]{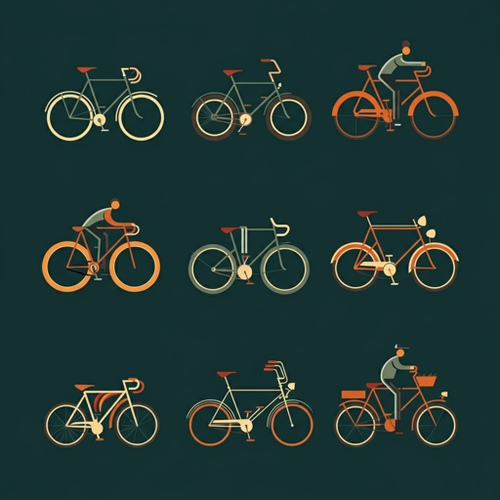} \includegraphics[width=.3\linewidth]{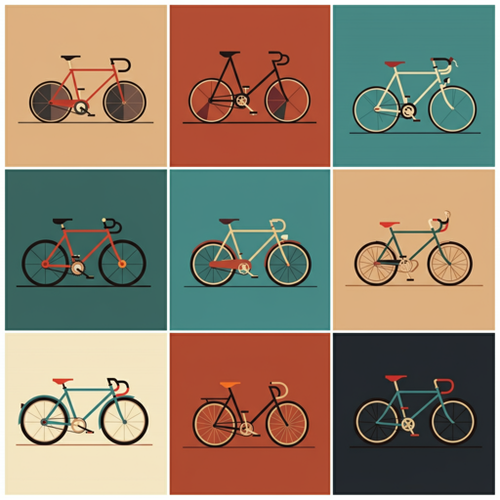} & 
\includegraphics[width=.3\linewidth]{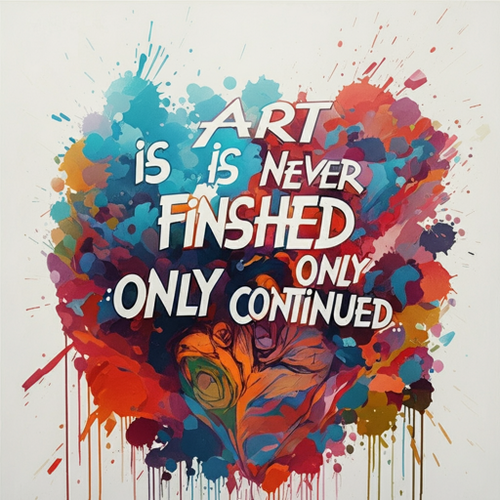} \includegraphics[width=.3\linewidth]{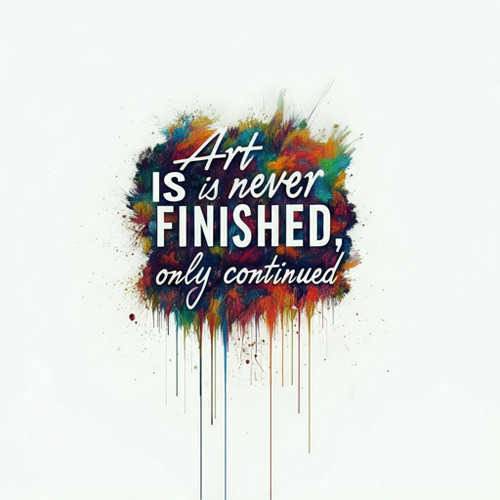} \includegraphics[width=.3\linewidth]{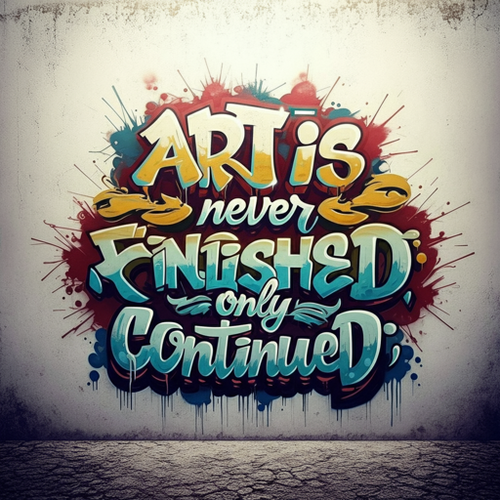}
\\ 
\makecell{7. \texttt{A white and blue bus sits on a brick ground. There is a white light} \\ \texttt{behind the bus. There is a person in a brown jacket standing} \\ \texttt{on the other side of the bus}} & \makecell{6. \texttt{beautiful isometric word 'DRAW' entirely made of pencils soft} \\ \texttt{smooth lighting, pastel colors,  trending on polycount} \\ \texttt{modular constructivism, blue background, physically based rendering, centered.}}\\

\includegraphics[width=.3\linewidth]{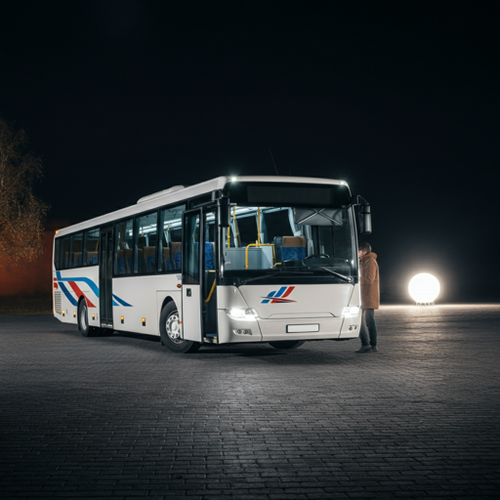} \includegraphics[width=.3\linewidth]{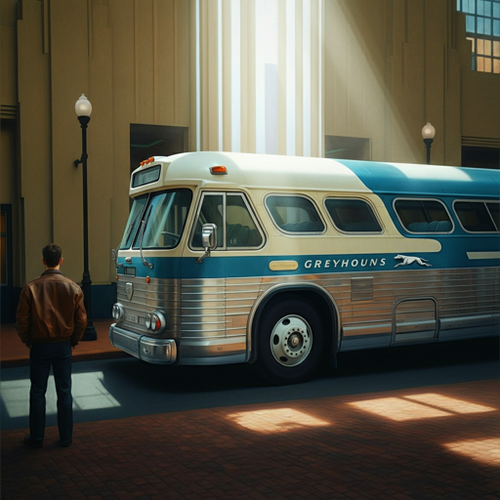} \includegraphics[width=.3\linewidth]{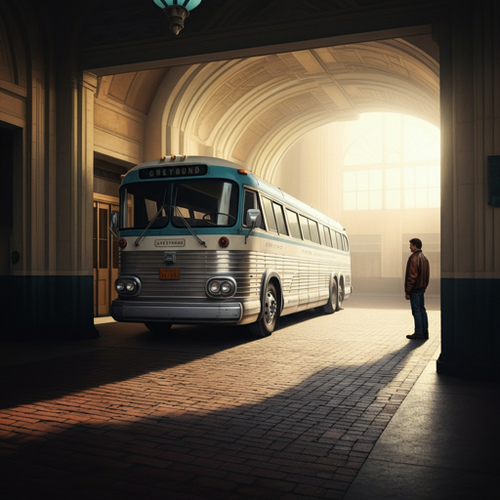} &
\includegraphics[width=.3\linewidth]{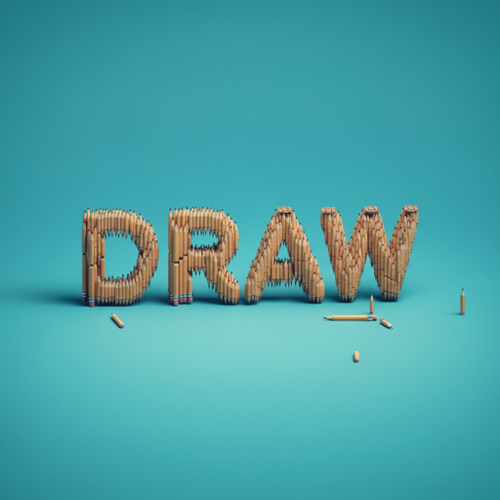} \includegraphics[width=.3\linewidth]{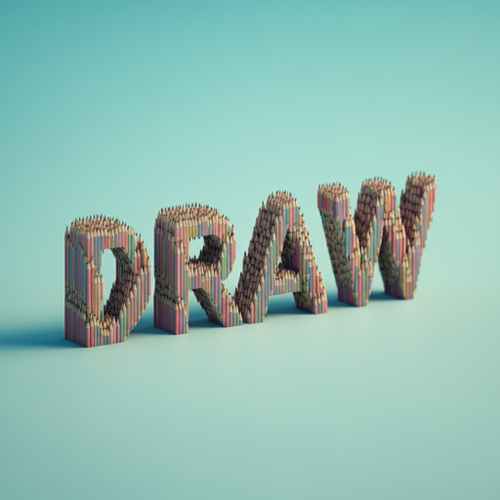} \includegraphics[width=.3\linewidth]{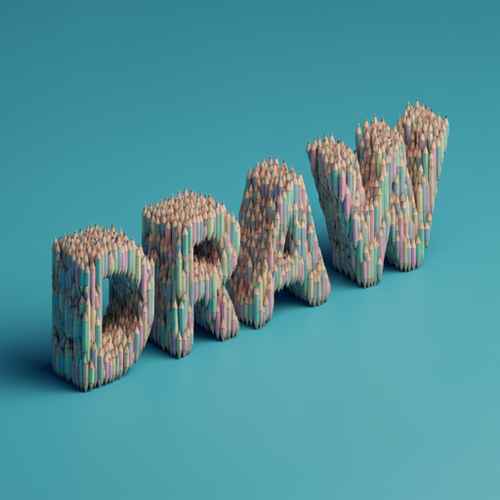}
\\
\bottomrule
\end{tabular}
\caption{Additional qualitative examples of the image trajectories generated by \ours. Refer to Table~\ref{tab:qualitative} for additional explanations.}
\label{tab:additional_qualitative}
}
\vspace{-2mm}
\end{table}

\end{document}